\documentclass[11pt]{article}

\usepackage{arabtex}

\usepackage{coling2018}

\usepackage{url}
\usepackage{float}
\usepackage{graphicx,subcaption}
\usepackage{smartdiagram}
\usepackage{pgf-pie}
\usepackage{caption} 
\usepackage[stable]{footmisc}

\usepackage{cp1256,utf8}
\setcode{utf8}

\date{}

\begin{document}

\title{Where Are You From? Let Me Guess!\\ 
	Subdialect Recognition of Speeches in Sorani Kurdish}


\author{
	\begin{tabular}[t]{c}
		Sana Isam and Hossein Hassani\\
		\textnormal{University of Kurdistan Hewl\^er}\\
		\textnormal{Kurdistan Region - Iraq}\\
		{\tt {\{sana.isam, hosseinh}\}@ukh.edu.krd}
	\end{tabular}
}

\maketitle

\begin{abstract}
Classifying Sorani Kurdish subdialects poses a challenge due to the need for publicly available datasets or reliable resources like social media or websites for data collection. We conducted field visits to various cities and villages to address this issue, connecting with native speakers from different age groups, genders, academic backgrounds, and professions. We recorded their voices while engaging in conversations covering diverse topics such as lifestyle, background history, hobbies, interests, vacations, and life lessons. The target area of the research was the Kurdistan Region of Iraq. As a result, we accumulated 29 hours, 16 minutes, and 40 seconds of audio recordings from 107 interviews, constituting an unbalanced dataset encompassing six subdialects. Subsequently, we adapted three deep learning models: ANN, CNN, and RNN-LSTM. We explored various configurations, including different track durations, dataset splitting, and imbalanced dataset handling techniques such as oversampling and undersampling. Two hundred and twenty-five(225) experiments were conducted, and the outcomes were evaluated. The results indicated that the RNN-LSTM outperforms the other methods by achieving an accuracy of 96\%. CNN achieved an accuracy of 93\%, and ANN  75\%. All three models demonstrated improved performance when applied to balanced datasets, primarily when we followed the oversampling approach. Future studies can explore additional future research directions to include other Kurdish dialects. 
\end{abstract}

\section{Introduction}
\label{sec:intro}

Understanding linguistic variety and how it affects communication depends heavily on dialects and subdialects. Kurdish provides considerable hurdles in natural language processing and language classifications due to its macro morphological structure and wide variety, as will be stated in detail. There are several dialects and subdialects of the Kurdish language and even within a subdialect, there might be differences across cities and towns.

The second main Kurdish dialect is Central Kurdish,  well-known as the Sorani dialect, which is spoken largely in the North-Eastern part of Iraq and the western part of Iran \newcite{hassanpour1992nationalism}. It is spoken in Sulaimani province, apart from Mergasour district; it is spoken in all areas of Erbil Province, Kirkuk and Halabja provinces in the Kurdistan Region of Iraq (KRI) and some other areas in Iraq, such as Basra and Jasan in Diyala province. It can also be spoken in several Kurdish cities in Iran, such as Mahabad, Bokan, Piranshahr, Sardsht, 
Shno (Ushnawiye), Naqade, Takab Jwanro, Rawansar, Salasi, Babajani and so forth \newcite{jabraelcritical2019}. In this respect, it has 
appeared as the official language of the Kurdistan Regional 
Government, parliament and other foundations \newcite{jabraelcritical2019}.
We aim to develop a dataset and create a model that can accurately detect and classify the Kurdish language's subdialects. Creating an audio dataset is critical for training and assessing machine learning models for subdialect categorisation. It enables us to correctly differentiate and categorise diverse subdialects by capturing their distinct phonetic and auditory properties.

The collection of a comprehensive voice dataset, consisting of 29 hours, 16 minutes and 40 seconds of recordings, encompassing six Kurdish Sorani subdialects (Garmiani, Hewleri, Karkuki, Pishdari, Sulaimani, Khoshnawi), represents a significant milestone for both us and other developers and computational linguists. This dataset will serve as a valuable resource for conducting research and delving deeper into the study of subdialects.

\subsection{The Kurdish Language}
More than 30 million people worldwide communicate with one another using the Indo-European language known as Kurdish. People are distributed in regions of different countries, primarily Iraq, Iran, Syria and Turkey, as well as Armenia, Azerbaijan and Georgia. Kurdish is a language with a variety of dialects, on which many studies have been done on both the Kurdish language and its various dialects. Many books and articles were written concerning this topic. This has been the subject of discussion among orientalists with certain viewpoints \cite{hanani2020spoken}.  

There is no consensus among researchers or linguists regarding the division of Kurdish dialects and subdialects. Most of them are divided into four dialects but with different names for them \cite{hassani2016automatic}. Table \ref{tab:subdialects} shows various views on the subdialects of Sorani Kurdish.

\begin{figure}
	\centering
	\includegraphics[width=\linewidth, height=15cm]{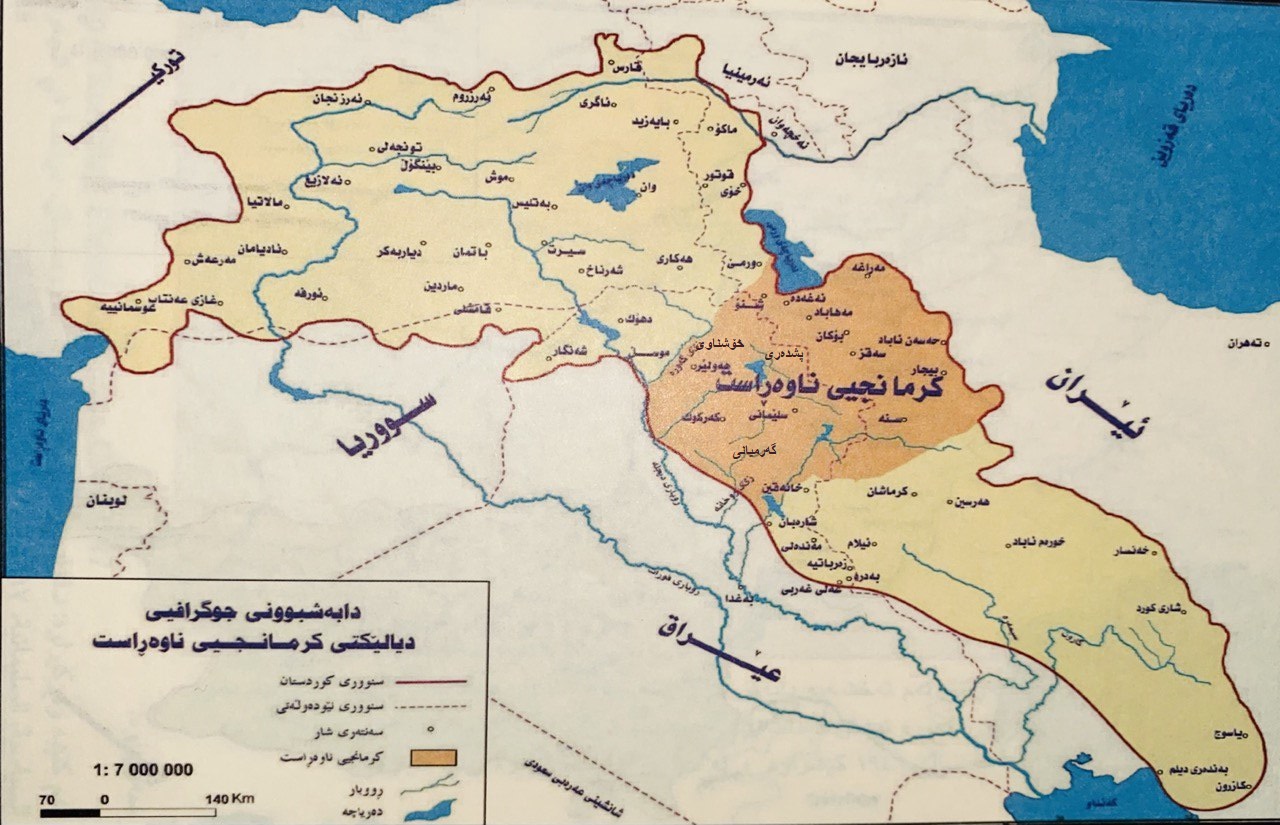}
	\caption[\fontsize{10}{12}Geographical distribution of Sorani dialect and its subdialects]{Geographical distribution of Sorani dialect and its subdialects, adopted from \newcite{zmanykwrdy2018}.}
	\label{fig:map}
\end{figure}

	\begin{table}
	\begin{center}
	\begin{tabular}{|p{3cm}|p{9cm}|}				
		\hline
		Source & Subdialect Names \\
		\hline
		\newcite{kurdishlanguagefuad83} & 
		{Mukri, Sorani, Ardalani, Sulaimani, Garmiani}
		\\
		\hline
		Tafiq Wahby  & {Mahabadi, Hewleri, Karkuki,  Mukri, Sorani, Ardalani, Sulaimani, Garmiani}
		\\
		\hline
		Dr. Jamal Nabaz  & {Karkuki, Mahabadi, Hewleri, Karkuki, Sulaimani,  Mukri, Sorani, Ardalani, Sulaimani, Garmian}\\
		\hline
		\newcite{izady2015kurds} & {Mukri, Ardalani, Garmiani, Khoshnaw, Pishdari, Warmawa, Kirmanshahi, Hewleri}
		\\
		\hline
		\end{tabular}
	\caption{Classification of Sorani subdialects} 
	\label{tab:subdialects}
	\end{center}
	\end{table}
Geographically, central Kurmanji is located in Iraq and Iran. A line separates central Kurmanji from the north Kurmanji up to Sirwan River and the high road between Khanaqen (in Iraq), Qasri Shirin, Kermanshah and Malayer (in Iran) and from the east of Hamrin Hills in the west (Iraq) up to the line extending in Sahand Mountain, Masirabad, Bijar and Asadabad in the East (Iran) \cite{kurdishlanguagefuad83}.
In addition, for a clearer understanding of the geographical distribution of the Sorani dialect and its subdialects, we adapted the map provided by \newcite{zmanykwrdy2018}. We have added three subdialects, namely Pishdari, Garmiani and Khoshnawi, and placed them in their respective approximate locations, illustrated in Figure \ref{fig:map}.
In Sorani, Arabic has influenced the subdialects of Iraq and Persian has influenced the subdialects of Iran. Some words are imported from Arabic into the dialects and subdialects of Kurdish located in Iraq \cite{kurdishlanguagefuad83}. 

Our investigation focused on the following Sorani subdialects:
	\begin{enumerate}
		\item \textbf{Sulaimani}- It is a subdialect spoken by the people of Sulaimaniah, Sulaimaniah, as a centre of the present district, has only existed for 239 years. Sulaiman Baban founded it in 1784 \cite{thereisnokurdishart}, located in northern Iraq and southern Kurdistan \cite{notessoane}.
		\item \textbf{Karkuki}- Along with other dialects like Hewleri, Sulaimanih and Snaye that are also present there, Karkuk also has its own subdialect \cite{SarkawtKarkuk2011}.
		Kakrkuki is identified as a subdialect of Central Kurmanji \cite{SarkawtKarkuk2011}. Due to its rich oil reserves, the region holds significant economic value and the Kurdish population of Karkuk has faced forced displacement on multiple occasions. As an accent, it's close to Garmiani. Karkuk is one of the large cities in the Kurdistan region of Iraq, between the two rivers of Serwan and Zei Bchuk (Lesser Zab) \cite{SarkawtKarkuk2011}. In Karkuk, seven branches of the subdialect exist Rozhbaiany, Kakeye, Shwany Kishk, Ajemye, Zengene, Sia Mensoury and Shekhany.
		\item \textbf{Hewleri}-	Hewleri is another subdialect of Sorani dialect, named after the Hewler city. The subdialect located in Hewler district in Iraqi Kurdistan (except Zebari province) \cite{kurdishlanguagefuad83}.
		\item \textbf{Khoshnawi}- The Khoshnawi subdialect is characterised by its significant scope and is primarily spoken by the inhabitants of Shaqlawa, Balisan and the surrounding villages \cite{KurdistanWria2009}. Khoshnawi subdialect geographically starts from Malakan in the north and extends southward until Gomespan. In the east, it begins from Serwchawe and stretches westward until Mela Nevyean. The subdialect encompasses Safeen Mountain and Shaqlawe. It is divided between the provinces of Hewler and Sulaimani. While it is a clan, as previously indicated, it identifies as a separate Sorani subdialect in specific sources.
		\item \textbf{Garmiani}-	This dialect is located in the western Sulaimaniah region and is primarily found in villages such as Kalar, Kifri, QaraTappeh and Tuz \cite{kurdishlanguagefuad83}.
		\item \textbf{Pishdari}-	Another distinctive Sorani Kurdish subdialect known as Pishdari is frequently used as a synonym for Qaladzaye.
		The subdialect is located in the northern region of Sulaimani, approximately 175 km away. Its center is in Qaladza, surrounded by various mountain chains, including Asos, Kurees, Doopeze, Bilfet, Mamend, Qendil, Zerine Kew, Pirane Resh and Kewe Resh. To the north, it borders Iran and Soran. To the west, it borders Rania, to the south, it borders Bingird, and to the east, it borders Iran. Due to the population's forced migrations, their accents have undergone changes over time.
	\end{enumerate}

There may be little or insufficient documentation, books or e-books that go into depth on the subdialects of Sorani Kurdish and how they're employed. Instead of digging into the nuances of dialects and subdialects, the materials that are now accessible mostly describe the physical locations of these cities.
	
The rest of the paper is organized as follows. Section~\ref{Sec-Work} reviews the literature and the related work. Section~\ref{Sec-Method} presents the method that the research follows. We provide the results and discuss the outcome in Section~\ref{sec-erd}. Finally, Section~\ref{Sec-Conc} concludes the paper and provides some ideas about future work.

\section{Related work}
\label{Sec-Work}
Numerous studies have been done on speech recognition, particularly in the areas of dialect and subdialect recognition. In speech, there are typically two distinct kinds of models, which are traditional models and deep learning models  \cite{ganapathiraju2004}.
The research on Kurdish speech processing is quite limited \cite{amani2021kurdish}. \newcite{hassani2016automatic} usded SVM model used for Kurdish dialect recognition in texts.

Regarding Kurdish recognition systems, the most recently developed Kurdish (Sorani) speech recognition system used Kaldi ASR. It used a trigram statistical language model and several acoustic models on diverse experiments, such as Tri1 used MFCC, delta, and delta-delta features for triphone modeling using the HMM-GMM algorithm. Tri2: HMM GMM-based tri-phone modeling with LDA and MLLT-transformed MFCC features Tri3: Triphone modeling based on HMM-GMM with MFCC, delta, delta-delta and SAT features, SGMM; Subspace Gaussian Mixture Model and with applying LDA to the MFCC features, Mono: Mono-phone modeling based on HMM with GMM and MFCC features They experienced this on the Jira dataset (the first Kurdish speech corpus-diphones based), designed by AsoSoft research and business group on natural language processing \cite{veisi2022jira}, Jira corpus is a collection of speech in the office using a pre-defined noise-free microphone and crowdsourcing in the Telegram social network using a smartphone microphone, then manually eliminating noise. 100 sentences for testing and 700 sentences for training over 11 topics were included, whereby 576 speakers made more than 42,000 tracks \cite{ortu2015jira}.

Based on SVM modeling of the N-gram of phones \newcite{richardson2009} explored a new class of methods. The SVM techniques offer comprehensible phone strings that represent a dialect. This methodology may be used to enhance existing dialect cue inventories from linguistics and fill part of the gap between automated approaches and linguistic analysis.

\subsection{Traditional Approaches of Speech Recognition}
\label{Traditional approaches}	
Traditional models that have been used are: Support Vector Machine (SVM), Naïve Bayes, Sequential Minimal Optimisation(SMO), C4.5 Decision Tree Classifier (J48), Zero Rule(ZeroR), Repeated Incremental Pruning to Produce Error Reduction (JRip) and bottleneck \cite{alshutayri2016arabic}.

\newcite{richardson2009} combined two spectral methods, SVM and N-gram, classify phone strings that represent a dialect. They use features such as the frequency centroid and standard deviation. Also,  \newcite{ziedan2016unified} proposed another adapted I-vector, as concept that originally designed for speaker recognition, for Arabic dialect recognition. 

Regarding the process, input speech is first turned into a sequence or lattice of tokens \cite{murhaf2013erg}, and then the N-gram is done. The outcome is utilised to predict the class labels for SVM sequence kernels. The approach was previously used and achieved success in \newcite{richardson2009} and \newcite{murhaf2013erg}, using their cases as dialects or languages. Some difficulties that they faced in the research can be described as the small amount of training data for all three dialects of English, Mandarin and Arabic, which makes understanding N-gram analysis challenging to evaluate whether the top characteristics are discriminative between the sets of speaker-dependent characteristics or discriminative between the specific dialects. Although an issue that complicated the analysis of their system to some extent was when the data for a specific dialect could have some dialect-related channel artifacts.

\newcite{alshutayri2016arabic} experimented with Naïve Bayes, SMO, J48, ZeroR and JRip classifiers and observed that by utilising SMO, they properly identified 6803 utterances and incorrectly classified 816 utterances, which achieved the best accuracy among other classifiers. After being verified and categorised by three human Arabic linguistic specialists, it was discovered that most of the misclassified utterances might have been better classified by converting them to Buckwalter to normal readable Arabic scripts, as reading the Buckwalter text is tough even for the experts on the Buckwalter transliteration system. Their method achieved an accuracy of around 50\% with a training set percentage split of 60:40 and was better than splitting the training set 80\%-90\% with an accuracy equal to 42.85\%. Besides the limitations related to writing style and the lack of official writing standards of dialects, they faced difficulties while collecting their datasets. Because they have both phonetic and acoustic information, traditional bottleneck features became attractive as a replacement for speech tasks such as Automatic Speech Recognition (ASR), Speaker Identification (SID) and  Language Identification (LID) after  Mel-Frequency Cepstral Coefficients (MFCC). Nonetheless, there are two possible negative effects.

\newcite{zhang2017dialect} proposed a new method for extracting BNF without relying on a transcribed corpus. This approach uses an unsupervised extraction diagram trained with estimated phonetic labels. The method was evaluated on Chinese dialect and Pan-Arabic datasets and consistently outperformed the baseline MFCC-based system. The proposed BNF achieved a relative improvement of +48\% in Equal Error Rate (EER) and +52\% in overall performance compared to the baseline. Even with limited training data, the proposed feature showed a relative improvement of up to 24\% without the need for a secondary transcribed corpus.

\newcite{salameh2018fine} proved that for classification jobs using discrete features, the MNB classifier is effective. As it identified the precision of a speaker with an accuracy of 67.9 percent for sentences with an average length of 7 words and above 90\% when taking 16 words into consideration, the process was using 3 datasets: Corpus-6 and Corpus-26 and a custom extracted from Twitter with 16,385 utterances in 49h36m.

\newcite{shafieian2022hidden} presented a practical method for Persian speech recognition using HMMs. They employed syllables as units for their HMM-based approach and incorporated features such as MFCC and PARCOR. The training was conducted on the FARSDAT dataset, which consists of two speech corpora: "Large FARSDAT" and "Small FARSDAT." The latter is a smaller Persian corpus recorded at low noise levels and includes phoneme-level segmentation and labelling. The speakers represent ten different Persian dialects, as a result, the HMM achieved a Word Error Rate (WER) of 18.3\%, and the system performance was improved by approximately 16\% through post-processing techniques. Despite various restrictions and challenges, it is widely recognised that the Persian language has relatively fewer computational studies compared to other languages. However, Persian stands out as a language with a rich vocabulary, allowing for the creation of numerous words through the addition of prefixes and suffixes \cite{shafieian2022hidden}.

Then some open-source programmes were developed and later used for speech recognition purposes. The Kaldi toolkit, a free open-source programme designed to a large extent by Daniel Povey, includes a basic library of Kaldi's C++ code that includes modeling of acoustic systems utilising subspace Gaussian mixture models (SGMM) and normal Gaussian mixture models, as well as all frequently used linear and affine transformations. \newcite{zeinali2019multi} uses the Kaldi toolkit for the first public large-scale speaker verification corpus in Persian. The DeepMine corpus contains more than 1850 annotators and 540,000 tracks, totalling more than 480 hours of speech. In addition, it showed that the DeepMine repository is more difficult to use than RedDots Speech Recognition Challenge 2015 (RSR2015) and RedDots, according to text-dependent findings. The model was trained on the DeepMine database with 5.9 hours  as a test set and 28.5 hours  as a large test set, their WERs were 4.44 and 4.09, respectively.

Another open-source system was used for speaker-independent Urdu speech recognition based on the HMM approach that was proposed for developing in \newcite{ashraf2010speaker}, the pen source framework called Sphinx4. They reported that their research is achieving satisfactory results in medium and large vocabulary sizes and used a small-sized vocabulary, specifically 52 isolated most spoken Urdu words. They received poor reliability from the system after utilising English acoustic models, as Sphinx4 libraries were used for some Latin languages like Italian or French and English. For all that, the WER was 60\%, as some Roman letters do not exist in English. The researchers solved this by recording ten samples from ten different narrators, then combining each file from 52-word files into a single-word file. This took a considerable amount of time, while more different phonemes in words decrease accuracy, while the same types of phonemes in words could confuse the system's recognition. Increasing the number of words will increase accuracy. The mean WER was 10.66\%.

Two components of a system for Arabic dialect detection were proposed in \newcite{ziedan2016unified} and are based on phonetic features and acoustic features. The first component is based on a phonotactic representation of the speech, and the second component oversees speech signal analysis to extract acoustic features. The choice is made through score-level fusion between the phonetic and acoustic systems following all testing and model phases. In this model, the PER with GMM-UBM model and the identity vector (i-vector) classifier were utilised. It could detect Egyptian Arabic dialects,  Levantine Arabic accent or dialect, Saud, Levantine Arabic dialect  and Gulf accents and their sub-dialects, which in total contain 3840 tracks. The model has been trained on the SARA (Spoken Arabic Regional Archive) dataset. The master dataset SARA consists of Arabic dialects that are spontaneous and canonical and gender independent, meaning that the speech was gathered through following specific instructions like reading and from human-human and human-machine conversations, as well as from the real world. When compared to other state-of-the-art systems, the acoustic feature might improve Arabic dialects more.

\subsection{Deep Learning Approaches of Speech Recognition}
\label{Deep learning pproaches}         

\newcite{hanani2020spoken} used the VarDial 2017 shared tasks to train and test their ADI systems, and they had high performance (68.7\%) on the x-vectors technique in their study. Fusing the model with i-vectors marginally enhanced its performance. In addition to MSA, the technique was used to distinguish between the five main dialects of Arabic-Gulf, Iraqi, Levantine, Egyptian, Meghrebi, Yemenite and Maltese—as well as some of its subdialects. Nevertheless, the model has several drawbacks. For example, DNN is not helpful for ASR due to its high computational requirements, which were overcome by employing multiple GPU cores to use DNN for speech recognition.

\newcite{li2018multi} proposed LAS (Listen, Attend and Spell), a neural network model that outputs a word sequence from an audio signal without the need for individual acoustic models, pronunciation models, language models, HMMs, etc. Sequence-to-sequence (Seq2Seq) learning model framework with attention is the foundation of the LAS model \cite{chorowski2015attention, sutskever2014sequence, chan2016listen}. Because traditional automated speech recognition systems require a distinctive pronunciation and language model for every dialect from the multi-dialect acoustic model, \newcite{li2018multi} designed a universal multi-dialect model for them in which mistakes are transmitted to the Language Model (LM) and Pronunciation Model (PM) if the Acoustic Model (AM) predicts a false collection of sub-word units from the incorrect dialect. The approach has several desirable features, including an improvement in low-resource languages and simplicity.

\newcite{veisi2020persian} discovered that using DNN for feature extraction is more effective in comparison with a shallow network and that using spectrogram features improves outcomes in comparison to MFCC features. For the first time, they combined Deep Bidirectional Long Short Term (DBLSTM) and Deep Belief Network (DBN) with the output layer Connectionist Temporal Classification (CTC) to create an AM, they increased system accuracy by using the bidirectional network rather than unidirectional model. By using Kaldi-DNN and HMM shows that using DBLSTM improved Persian phoneme recognition accuracy, with DBLSTM neural network the LSTM, DLSTM and BLSTM were used too.

\newcite{gultekin13turkish} suggested a phoneme-based RNN-LSTM language model instead of the n-gram model in PPRLM since RNN and n-gram models achieve less accuracy than LSTM networks. The reason that they didn't compare it to other studies, was as it is the only study to use the LSTM language model in PPRLM for dialect recognition since 2020. They trained models with the ends of sentences and showed it works better than training with the entire sentence, then found out that classifying dialects can be done only by looking at the sentence endings. In this new investigation, it was suggested only for the audio of 1 second and 0.5 seconds and achieves (83.8- 84.2\%). Therefore, for the long sentences (3 seconds), the whole sentence was used, and the accuracy was 84.4\%. This supports the previous studies that, with increasing the test duration, the accuracy would increase too. The model could successfully recognise the four region dialects: Ankara, Trabzon, Alanya and Kibris. The whole dataset contains 2.7 hours of noise-free audio. The audio was recorded from older people with low levels of education, not on purpose but due to the characteristics of the people selected. They could only find those. In the end, they suggested they could improve their model by using BLSTM.

\newcite{farooq2019improving} investigated and compared three different state-of-the-art models for the Urdu LVCSR system: 3-gram LM, RNNLM and Text normalised acoustic model + RNNLM, then TDNN-BLSTM was chosen to develop the system, and recording the decoded output lattice was done using RNNLM. WER was 13.5 when they developed their Urdu corpora, which contains 300 hours of noise-free recordings from 1671 speakers with a vocabulary of 199,000 words. The models were 3-gram LM, RNNLM and Text Normalised Acoustic Model + RNNLM, and the WER was 18.64, 16.94 and 13.50, respectively. The investigation was constrained by the fact that some Urdu words might be written in two alternative ways, if the decoded version differs from the reference text, ASR is penalised as one replacement, as well as some words that ASR occasionally inserts spaces into, can be true with or without spaces, yet they are incorrect for WER computation. After text normalisation of the training and test sets, these penalties can be eliminated by retraining transcriptions.

The theory of CNN and its practical application techniques are now evolving in tandem with significant growth in the number of CNN layers, which raises the computational complexity of the systems that use them, the network architecture, for instance,            \cite{valueva2020application}. Based on the shared-weight design of the convolution kernels or filters that slide along input features and produce translation-equivalent responses known as feature maps, CNNs are also known as Space-Invariant Artificial Neural Networks (SIANN) \cite{zhang1990parallel}. 

To summarize, traditional models such as SVM, Naïve Bayes and HMMs have been used for dialect recognition and speech processing. N-gram analysis and feature adaptations like I-vector and bottleneck have shown promising results. Challenges include limited training data and artifacts. Open-source programmes like Kaldi toolkit aid in acoustic system modeling. Ongoing efforts aim to improve accuracy and performance in these areas. While deep learning approaches have gained prominence in recent years. Deep learning models such as CNNs, RNNs and transformer models have shown promising results in these tasks. These models can automatically learn hierarchical features and capture complex patterns in speech data. However, deep learning approaches require large amounts of labelled data and computational resources for training. Nonetheless, they offer the potential for further advancements in dialect recognition and speech processing.The usage of these techniques in music genre classification has shown a promising results \newcite{zuhair2021comparing,chettiar2021music} that could be replicated in speech classification as too. We intend to base our experimental approaches on those findings.

\section{Method}
\label{Sec-Method}
The following section describes the method that the research follows.
  
\subsection{Data Collection}
The theme of our speech corpus is a normal daily conversation about the personal background of speakers, which is obtained using an interview guided by lead questions  in a way that the speech includes proper names, numerals, dates and times, the speakers' past lives, their education, and such. The interview guide includes various sections: The participant's background, routine duties, previous experiences, hobbies and interests, long answers to let the speakers talk about their weekends, vacations, life lessons and personal stories.

To increase the number of participants, we choose a wide variety of questions for different groups of them. We assume that the interests, perspectives and experiences of participants are vary depending on their ages, professions, occupations, and educational backgrounds. For instance, elderly participants who have long experiences in life may have insightful stories to tell about their memories, experiences or life lessons. Younger people, on the other hand, could be more likely to talk about their goals, favorite books, movies or online hobbies. We also consider the individuals' particular roles and jobs. For example, farmers or shepherds receive questions specifically geared toward their experiences, allowing them to share their perceptions of their daily activities, professional difficulties or relationships with the natural world. Similarly, a PhD. holder receives questions that aim to get answers that are more about their specific expertise, areas of interest in research or professional backgrounds. 

On the same vein, various roles and lifestyles of the participants, such as those of housewives and college students, can affect the talks. Whether it is their academic endeavors, extracurricular activities or their responsibilities and experiences inside the home, some questions target each of the mentioned sectors. We prefer that every participant, regardless of age, career, background or occupation, finds relevance and engagement in the conversation by incorporating a wide range of questions. This strategy not only makes the Conversational scenario inclusive but also gives a better and more comprehensive view of the participants' lives, interests and viewpoints.

Additionally, some questions cover subjects such as friendships, vacations, and place of birth, allowing all participants to contribute their viewpoints and experiences. Furthermore, to collect various speech constructs, the questions include a range of sentence structures, including the past, present, and continuous tenses. The questions also include positive and negative statements and various phrase intonations, including rising, falling and questioning tones. That allows participants to express themselves utilizing a variety of sentence structures and linguistic elements.

We plan the dialogue structure in a way that encourages rapport and desire for participants to engage in a more natural conversation. It should start with straightforward questions like number counting and basic personal information before moving on to subjects including interests, hobbies, daily routines and lifestyles. Finally, the conversation plan uses open-ended questions to collect detailed narratives and life stories, allowing participants to offer insightful commentary. This strategy makes recording diverse and rich voice data easier while ensuring a smooth flow of conversation.

\subsection{Speech Data Editing and Segmentation}
The speech recordings are subject to an editing procedure that removes long pauses, excessive background noise and any intervening voices, focusing exclusively on the intended speaker's voice. We use different time frames to evaluate the performance, accuracy, and error rates in our approaches.

\subsection{Data preprocessing}
The datasets are stored in the \textit{wav} format. This format is renowned for its uncompressed nature and superior sound quality in contrast to the MPEG audio Layer-3 (mp3) file formats.

\subsection{Feature Extraction}
We use the Mel Frequency Cepstral Coefficients (MFCC) for feature extraction. Figure \ref{fig:process-flow} outlines the process of extracting the MFCC.
\tikzset{
	startstop/.style={rectangle, rounded corners, minimum width=3cm, minimum height=1.5cm, text centered, draw=black, fill=red!30},
	process/.style={rectangle, minimum width=3cm, minimum height=1.5cm, text centered, draw=black, fill=blue!30},
	arrow/.style={thick,->,>=stealth},
	connector/.style={thick,->,>=stealth,rounded corners=2pt},
}

\begin{figure}[htbp]
	\centering
	\begin{tikzpicture}[node distance=4.5cm]
	\node (Framing) [startstop, fill=red!30] {\textcolor{black}{Framing}};
	\node (FFT) [process, right of=Framing, fill=blue!30, text width=3cm, align=center] {\textcolor{black}{Fast Fourier\\Transform}};
	\node (Mel) [process, right of=FFT, fill=green!30, text width=3cm, align=center] {\textcolor{black}{Mel-frequency\\Filtering}};
	\node (Log) [process, below of=Mel, fill=orange!30] {\textcolor{black}{Log Energy}};
	\node (DCT) [process, left of=Log, fill=purple!30, text width=3cm, align=center] {\textcolor{black}{Discrete Cosine\\Transform}};
	
	\node (Select) [process, left of=DCT, fill=yellow!30] {\textcolor{black}{Selecting Coefficients}};
	
	\draw [arrow] (Framing) -- (FFT);
	\draw [arrow] (FFT) -- (Mel);
	\draw [arrow] (Mel) -- (Log);
	\draw [arrow] (Log) -- (DCT);
	\draw [arrow] (DCT) -- (Select);
	
	\draw [thick, black] ($(Framing.north west)+(-0.6cm,0.5cm)$) rectangle ($(Select.south east)+(10cm,-0.5cm)$);
	\end{tikzpicture}
	\caption{Feature extraction workflow}
	\label{fig:process-flow}
\end{figure}
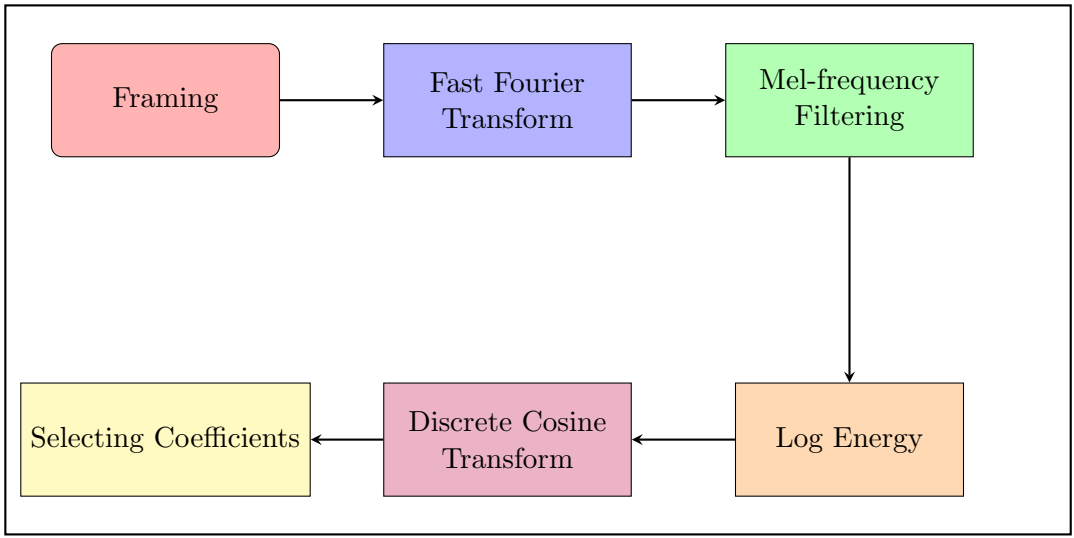

\subsection{Approaches}
We adapt two neural network approaches for our study: Artificial Neural Network (ANN) and Convolutional Neural Networks (CNN). The following sections describe the adaption of those approaches.

\subsubsection{Artificial Neural Network}

\begin{figure}
	\centering
	\begin{tikzpicture}[x=1.5cm, y=2.5cm, >=stealth]
	
	\path (-1.2, -2) rectangle (9.2, 2);
	
	\draw[thick, rounded corners, blue!50!black] (current bounding box.south west) rectangle (current bounding box.north east);
	
	\node[circle, draw=black, fill=gray!, minimum size=1.2cm] (input) at (0, 0) {\textcolor{blue!50!black}{Input}};
	
	\foreach \y in {1,...,3}
	\node[circle, draw=black, fill=gray!35, minimum size=1.2cm] (hidden1-\y) at (2, \y-2) {$\mathbf{ReLU}$};
	
	\foreach \y in {1,...,3}
	\node[circle, draw=black, fill=gray!30, minimum size=1.2cm] (hidden2-\y) at (4, \y-2) {$\mathbf{ReLU}$};
	
	\foreach \y in {1,...,3}
	\node[circle, draw=black, fill=gray!25, minimum size=1.2cm] (hidden3-\y) at (6, \y-2) {$\mathbf{ReLU}$};
	
	\foreach \y in {1,...,3}
	\node[circle, draw=black, fill=gray!20, minimum size=1.2cm] (output-\y) at (8, \y-2) {};
	
	\draw[->] (input) -- (hidden1-1);
	\draw[->] (input) -- (hidden1-2);
	\draw[->] (input) -- (hidden1-3);
	
	\foreach \i in {1,...,3}
	\foreach \j in {1,...,3}
	\draw[->] (hidden1-\i) -- (hidden2-\j);
	
	\foreach \i in {1,...,3}
	\foreach \j in {1,...,3}
	\draw[->] (hidden2-\i) -- (hidden3-\j);
	
	\foreach \i in {1,...,3}
	\foreach \j in {1,...,3}
	\draw[->] (hidden3-\i) -- (output-\j);
	
	\node[above, blue!50!black] at (0, 0.3) {{Input Layer}};
	\node[above, blue!50!black] at (2, 1.27) {{Hidden Layer1}};
	\node[above, blue!50!black] at (4, 1.27) {{Hidden Layer2}};
	\node[above, blue!50!black] at (6, 1.27) {{Hidden Layer3}};
	\node[above, blue!50!black] at (8, 1.27) {{Output Layer}};
	\node[below, blue!50!black] at (2, -1.3) {\textcolor{blue!50!black}{{.}}};
	\node[below, blue!50!black] at (4, -1.3) {\textcolor{blue!50!black}{{.}}};
	\node[below, blue!50!black] at (6, -1.3) {\textcolor{blue!50!black}{{.}}};
	\node[below, blue!50!black] at (8, -1.3) {\textcolor{blue!50!black}{{.}}};
	
	\node[below, blue!50!black] at (2, -1.5) {\textcolor{blue!50!black}{{.}}};
	\node[below, blue!50!black] at (4, -1.5) {\textcolor{blue!50!black}{{.}}};
	\node[below, blue!50!black] at (6, -1.5) {\textcolor{blue!50!black}{{.}}};
	\node[below, blue!50!black] at (8, -1.5) {\textcolor{blue!50!black}{{.}}};
	\node[below, blue!50!black] at (2, -1.7) {\textcolor{blue!50!black}{512}};
	\node[below, blue!50!black] at (4, -1.7) {\textcolor{blue!50!black}{256}};
	\node[below, blue!50!black] at (6, -1.7) {\textcolor{blue!50!black}{64}};
	\node[below, blue!50!black] at (8, -1.7) {\textcolor{blue!50!black}{6}};
	
	\node[above, blue!50!black] at (1, 0.7) {\textcolor{blue!50!black}{w(1)}};
	\node[above, blue!50!black] at (3, 1) {\textcolor{blue!50!black}{w(2)}};
	\node[above, blue!50!black] at (5, 1) {\textcolor{blue!50!black}{w(3)}};
	\node[above, blue!50!black] at (7, 1) {\textcolor{blue!50!black}{w(4)}};   
	\end{tikzpicture}
	\caption [ANN architecture] {ANN architecture for dialect classification with the input acoustic features of speech signal and subdialects as targets with ReLU activation function in hidden layers}
	\label{fig:neural_network}
\end{figure}
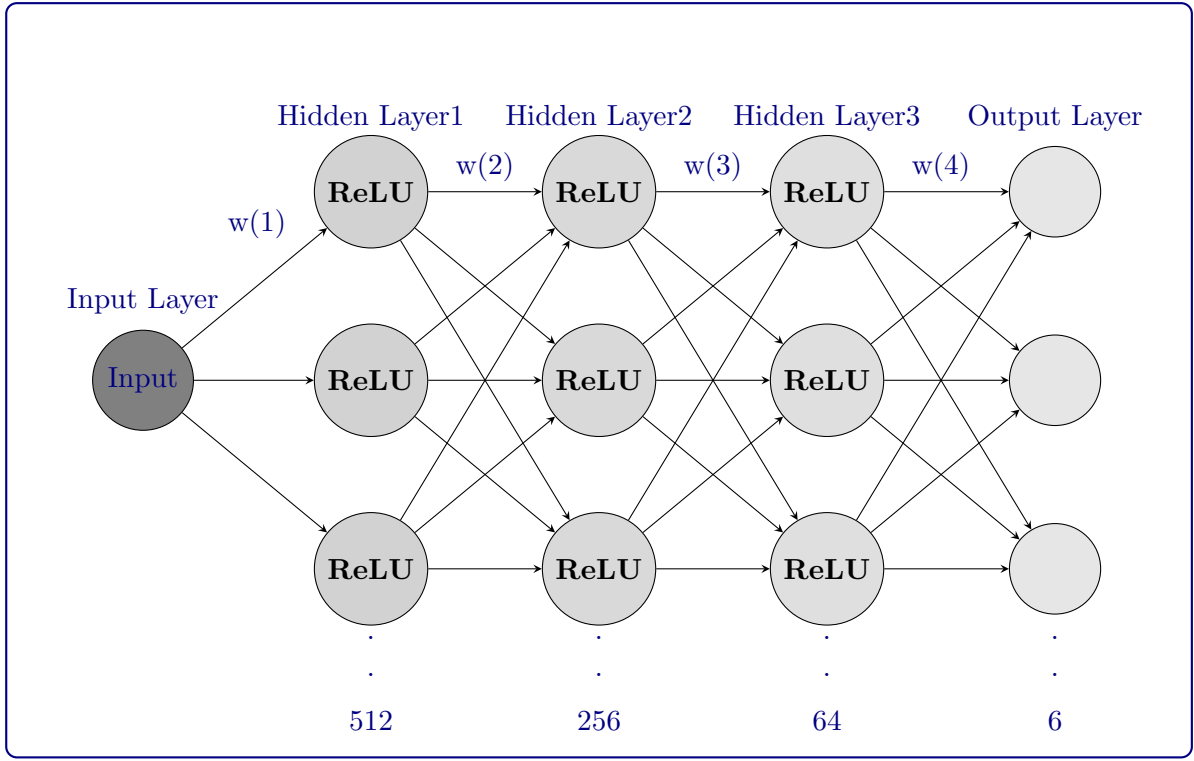

Figure~\ref{fig:neural_network} the adapted ANN with an input layer, three hidden layers, and the output layer. In the implementation, we utilize Keras and \textit{scikit-learn} libraries to train ANN.


The model consists of layers with varying numbers of nodes. The first layer is constructed with 512 nodes, while the second and third layers have 256 and 64 nodes, respectively. The non-linear activation function ReLU is used in all three hidden layers of the model. The first layer is the input and the last is the output. The number of nodes in the output layer is six, i.e., the number of subdialects.

\subsubsection{Convolutional Neural Networks}

%

\begin{figure}
\centering
\includegraphics[height=0.3\textheight, width=1\textwidth]{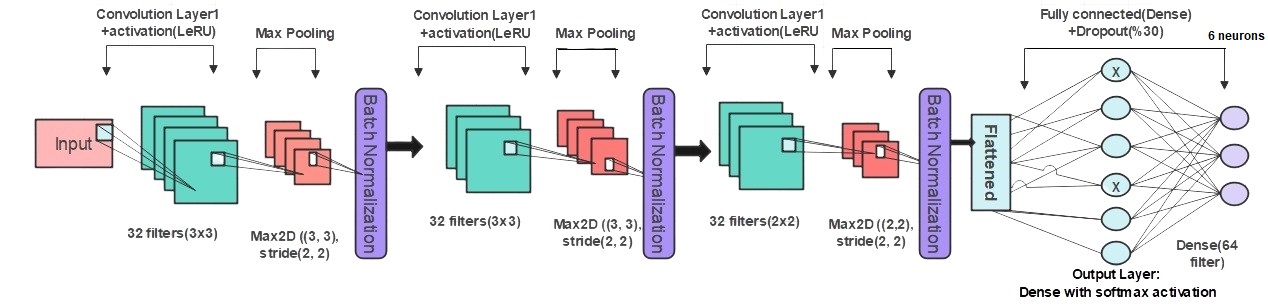}
\caption{The adapted CNN}
\label{fig:CNN}
\end{figure}

Figure~\ref{fig:CNN} shows the CNN model that comprises three subsequent convolutional layers, each followed by ReLU activation and max pooling. The initial convolutional layer implements downsampling of the feature maps by applying max pooling with a pool size of 3x3 and a stride of 2x2. The third layer of convolution in the model employs the max pooling technique with a pool size of 2x2 and stride. The utilisation of batch normalisation is implemented in the initial and secondary convolutional layers to normalise the activations of the layers, thereby enhancing the efficacy of the training process. Following the convolutional layers, the resultant output is transformed into a 1D (dimensional) vector using the \textit{Flatten()} layer. This step involves preparing the data for the fully connected layers that follow. The architecture comprises a densely connected layer of 64 neurons and ReLU activation. The dropout regularisation technique has been employed with a rate of 0.3 to address the overfitting issue. The ultimate layer of output consists of six neurons that utilise Softmax activation to generate the anticipated probabilities for the various categories.

To calculate the total number of neurons, we sum up the neuron. We have 288 neurons in the first and second layers and 128 neurons in the third layer. The flatten Layer does not add any additional neurons. It reshapes the output of the last convolutional layer to a 1-D vector. the First Dense Layer, which was fully connected, outputs 64 neurons and 6 neurons would be the output of the output layer based on the number of classes.
Therefore, the overall number of neurons in the model is 288 + 288 + 128 + 64 + 6 = 774.

In the next step, we calculate the candidate cell state (\(\widetilde{C_t}\)) using the hyperbolic tangent function (\textit{tanh}) (See Formula~\ref{frm:cnn}).

\begin{equation}
\widetilde{C_t} = \textit{tanh}(W_C \cdot [h_{t-1}, x_t] + b_C)
\label{frm:cnn}
\end{equation}

The result \(\widetilde{C_t}\) represents the candidate values could be added to the cell state in the LSTM unit.

\subsubsection{Recurrent Neural Networks-Long Short-Term Memory}

For Recurrent Neural Networks-Long Short-Term Memory(RNN-LSTM), we consider five different divisions of training, validation and testing: 50:25:25, 60:20:20, 70:15:15, 80:10:10, and 90:5:5. We build a model when two consecutive LSTM layers are used. The model consists of two stacked LSTM layers, where the first layer processes the input sequence and the second layer processes the output sequence of the first layer. We define the first LSTM layer with 64 units. It takes the input\_shape as its input and has \textit{return\_sequences=True}, which means it will return the output sequence rather than just the final output. This feature is typically used when stacking multiple LSTM layers. Furthermore, the second LSTM layer is also defined with 64 units. It does not have \textit{return\_sequences=True}, which means it will only return the final output instead of the entire sequence. Following the LSTM layers, a dense layer is added to the model using the Dense class from Keras. It consists of 64 units and uses the ReLU activation function.To prevent overfitting, a dropout layer is included after the dense layer. Finally, the Dense class with a Softmax activation function adds an output layer to the model. It consists of 6 units, corresponding to the number of classes in the classification task.

In addition, callbacks are defined to enhance the training process. In our implementation, the Earlystopping callback is used to monitor the validation loss and stop training early if there is no improvement in 10 epochs. The Earlystopping callback is passed, as we mentioned earlier. 
 
\section{Experiments, Results, and Discussion}
\label{sec-erd}

The following sections report on the data collection, describes the experiments, presents the results, and discusses the outcomes.

\subsection{Data Collection}

We designed a guideline according to what we mentioned in the methodology. The guideline included 83 questions in five sections: counting (two questions), biography (20), daily routine (25), hobbies and interests (26), and long answers (10).  

\subsubsection{ Speaker Identification}
In recruiting speakers we chose individuals proficient in the Kurdish-Sorani subdialects being targeted, free of any speech impediments and comfortable with having their speech recorded. The identification of speakers from specific subdialects presented difficulties, causing us to ask for the help of students in Hewl\^er (Erbil) who were native speakers of the specific Kurdish-Sorani subdialects and stayed in university accommodations. Following that, we visited the districts, towns, and cities where they reside, such as Balisan, Garmian, Sulaimani, and Kirkuk. In areas that we could not reach for various reasons, we recoreded the interviews through online platforms, such as WhatsApp and Telegram.

\subsubsection{Ethical Considerations}
Before starting the recording sessions, an individual proficient in the language in question was engaged to aid in selecting and assessing the speaker's accent. We also asked the participants to provide us permission by signing a formal document to authorize us to disseminate the records publicly. For the online participants, the act of signing the contract was facilitated by acquaintances or relatives of the parties involved, who granted permission to use the voice recordings.

\subsection{Recording Locations}
The recording sessions were conducted in a range of stable environments, such as libraries, reception areas of accommodations, the living rooms of participants and classrooms within academic institutions. We deliberately recorded speech in typical, natural settings in the above-mentioned places. Throughout the recording process, ambient sounds and background noise typically present in such settings were recorded, thus providing a natural acoustic background. The objective was to capture audio in settings that simulated typical, everyday circumstances, ensuring a balance between ambient noise and clarity of speech, avoiding excessively loud or completely noise-free environments.

\subsection{Recording Configuration }
During the recording sessions, the participants were asked to speak into a microphone connected to a laptop. The microphone was placed near the participants while we were at a distance. This setup allowed the recording of the participants' speech data to occur simultaneously. Regarding the hardware, the recording microphone met the following specifications: 192K/24b sample rate, low impedance output (680Q), 100Hz–18000Hz frequency range and a maximum input sound pressure level of 125 dB, connected to laptop HP Pavilion x360 Convertible 14-dh2xxx.
In addition, the recording and editing process involved taking advantage of Audacity, a software application deliberately selected for its capacity to manage and edit audio recordings effectively. To improve the quality and realism of the auditory experience, stereo channels were utilised instead of mono \cite{monovsstereo}. The audio recordings were obtained at a sampling frequency of 44100 Hz and sample rate of 22050 Hz and 11,025 Hz of bandwidth utilising a 32-bit depth and were saved in the.aup3 (Audacity 3 Project File) format using the Audacity software. Following the editing procedure, the data was encoded using Pulse Code Modulation (PCM) with a bit depth of 16. The resultant files were then stored in the wav (Waveform Audio File) format. The voices of the participants who submitted their recordings online were transmitted via platforms such as WhatsApp or Telegram and were subsequently converted into the .wav format.

\subsection{The Environment Configuration}
\begin{figure}
	\centering
	\includegraphics[width=12cm, height=7cm]
	{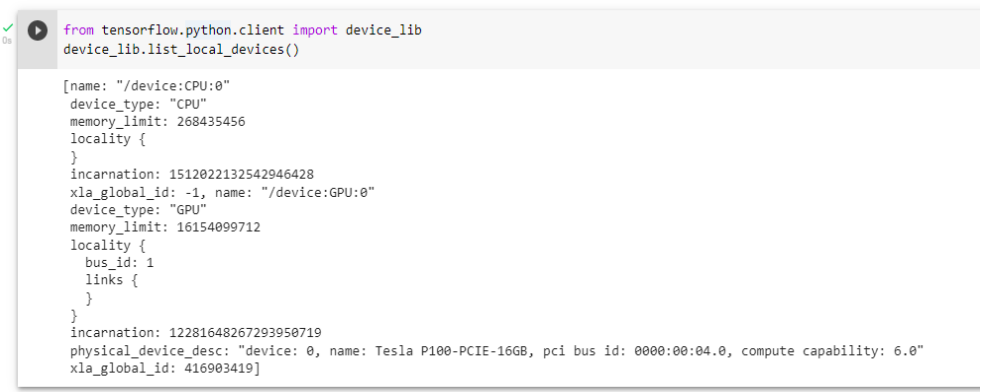}
	\caption{The type of GPU provided by Google Collaboratory.}
	\label{fig:colab}
\end{figure}

Python has been chosen as the programming language for model development because its numerous efficient libraries facilitate a more straightforward and faster process.
The libraries used for this thesis are as follows:
Regarding the environment, The utilisation of Google Collaboratory is preferred for this research due to the high computational requirements for training deep learning models, which are often challenging to access. The utilisation of a Jupiter notebook platform enables the facilitation of deep learning model training on Graphics Processing Units (GPUs) via cloud computing \cite{GarbadeColab}.
Additionally, Colab provides cost-effective subscription options such as Colab Pro and Colab Pro+, which offer enhanced features such as a higher-performing GPU, increased RAM and extended runtime. Despite the continued limitations on GPU time, it remains significantly higher than that of the free plan, as noted by \newcite{GarbadeColab}. As depicted in Figure \ref{fig:colab}, the NVIDIA Tesla P100 GPU is typically the provided option.
Regarding the platforms, we used Pycharm and Jupyter and the libraries were:
\begin{itemize}
	\item \textit{TensorFlow}: TensorFlow is a powerful open-source library for machine learning and deep learning. It provides a comprehensive set of tools and functionalities for building and training neural network models.
	\item \textit{TensorFlow.keras.callbacks.EarlyStopping}: This callback provided by TensorFlow Keras allows for early stopping during model training based on specified criteria.
	\item \textit{Json}: Used for reading and loading data from a JSON file.
	\item \textit{Numpy}: Used for array manipulation and processing.
	\item \textit{TensorFlow.Keras}: Used for building and training the neural network model, it was an independent library, but, starting from TensorFlow 2.0, Keras became integrated into TensorFlow as the official high-level API.
	\item \textit{Time}: Used for measuring the training time in seconds.
	\item \textit{Sklearn.model\_selection.train\_test\_split}: The data is split into training, validation and testing sets.
\end{itemize}
Regarding the hardware, the specifications mattered as some of the experiments were run on the laptop. The specifications were as follows: 
\begin{itemize}
	\item CPU: Intel(R) Core(TM) i7.
	\item RAM: 8.00 GB.
	\item Operating System: 64-bit.
	\item GPU: Intel(R) Iris (R) Plus Graphics.
\end{itemize}

\subsection{Dataset Preparation}
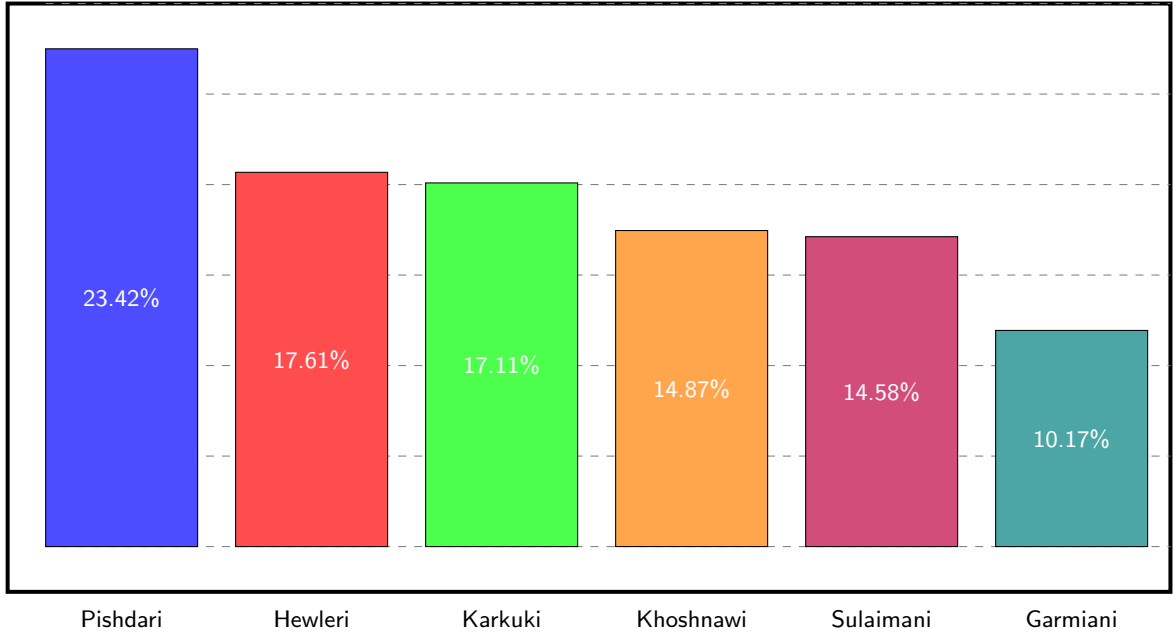
\begin{figure}
	\centering
	\begin{tikzpicture}[yscale=1.2,
	bar1/.style={fill=blue!70},
	bar2/.style={fill=red!70},
	bar3/.style={fill=green!70},
	bar4/.style={fill=orange!70},
	bar5/.style={fill=purple!70},
	bar6/.style={fill=teal!70},
	label/.style={font=\sffamily\small},
	value/.style={font=\sffamily\small, text=white},
	border/.style={line width=1.5pt},
	grid/.style={gray, dashed}
	]
	
	\draw[border] (-0.5,-0.5) rectangle (14.8,6);
	
	\foreach \y in {0,1,...,6}
	\draw[grid] (0,\y) -- (14.8,\y);
	
	\draw[bar1] (0,0) rectangle (2, {23.42/23.42*5.5}) node[value, above] at (1, {23.42/23.42*5.5}) {};
	\node[value] at (1, {23.42/23.42*5.5/2}) {23.42\%};
	
	\draw[bar2] (2.5,0) rectangle (4.5, {17.61/23.42*5.5}) node[value, above] at (3.5, {17.61/23.42*5.5}) {};
	\node[value] at (3.5, {17.61/23.42*5.5/2}) {17.61\%};
	
	\draw[bar3] (5,0) rectangle (7, {17.11/23.42*5.5}) node[value, above] at (6, {17.11/23.42*5.5}) {};
	\node[value] at (6, {17.11/23.42*5.5/2}) {17.11\%};
	
	\draw[bar4] (7.5,0) rectangle (9.5, {14.87/23.42*5.5}) node[value, above] at (8.5, {14.87/23.42*5.5}) {};
	\node[value] at (8.5, {14.87/23.42*5.5/2}) {14.87\%};
	
	\draw[bar5] (10,0) rectangle (12, {14.58/23.42*5.5}) node[value, above] at (11, {14.58/23.42*5.5}) {};
	\node[value] at (11, {14.58/23.42*5.5/2}) {14.58\%};
	
	\draw[bar6] (12.5,0) rectangle (14.5, {10.17/23.42*5.5}) node[value, above] at (13.5, {10.17/23.42*5.5}) {};
	\node[value] at (13.5, {10.17/23.42*5.5/2}) {10.17\%};
	
	\node[label, align=center] at (1, -0.8) {Pishdari};
	\node[label, align=center] at (3.5, -0.8) {Hewleri};
	\node[label, align=center] at (6, -0.8) {Karkuki};
	\node[label, align=center] at (8.5, -0.8) {Khoshnawi};
	\node[label, align=center] at (11, -0.8) {Sulaimani};
	\node[label, align=center] at (13.5, -0.8) {Garmiani};
	
	\end{tikzpicture}
	\caption{\fontsize{10}{12}Distribution and percentage of each subdialect in Sorani Nas}
	\label{subdialects_dostribution}
\end{figure}

The total record time is 29 hours, 16 minutes and 40 seconds. Figure \ref{subdialects_dostribution} and Table \ref{tab:recording_duration} illustrate the details of the dataset, which we named it Sorani Nas (in English, Sorani Recognizer).

\begin{table}
	\begin{center}
		\caption{Duration of recordings for different subdialects in Sorani Nas}
		\begin{tabular}{|p{3cm}|p{6cm}|}
			\hline
			{Subdialects} & \fontsize{12}{12} {Duration} 
			\\ \hline
			Garmiani & 2 hours, 58 minutes, 34 seconds \\ \hline
			Sulaimani & 4 hours, 29 minutes, 27 seconds \\ \hline
			Khoshnawi & 4 hours, 50 minutes, 22 seconds \\ \hline
			Karkuki & 5 hours, 45 minutes \\ \hline
			Hewleri & 5 hours, 13 minutes \\ \hline
			Pishdari & 6 hours, 49 minutes, 16 seconds \\ \hline
		\end{tabular}			
		\label{tab:recording_duration}
	\end{center}
\end{table}

\begin{figure}
	\centering
		\begin{subfigure}{0.45\textwidth}
			\centering
			\resizebox{\linewidth}{!}{%
				\begin{tikzpicture}
				\pie[radius=6, rotate=270, color={yellow!70!black, teal!70, orange!80!black, purple!40}, text=legend, every annotation/.style={font=\huge}]{5.83/Master and PhD, 48.54/Student, 24.27/Bachelor degree, 21.36/Uneducated}
				\end{tikzpicture}
			}
			\caption*{(a) Educational Level Distribution}
			\label{fig:educationdistribution}
		\end{subfigure}
		\hfill
		\begin{subfigure}{0.45\textwidth}
			\centering
			\resizebox{\linewidth}{!}{%
				\begin{tikzpicture}
				\pie[radius=6, rotate=270, color={yellow!70!black, green!40!black, blue!40}, text=legend, every annotation/.style={font=\huge}]{16.5/50-90 years old, 27.2/30-50 yo, 56.3/15-30 yo}
				\end{tikzpicture}
			}
			\caption*{(b) Age Distribution}
			\label{fig:agedistribution}
		\end{subfigure}
	\caption{The distribution of ages and educational level of the participants in Sorani Nas dataset}
	\label{fig:ageeducationdistribution}
\end{figure}
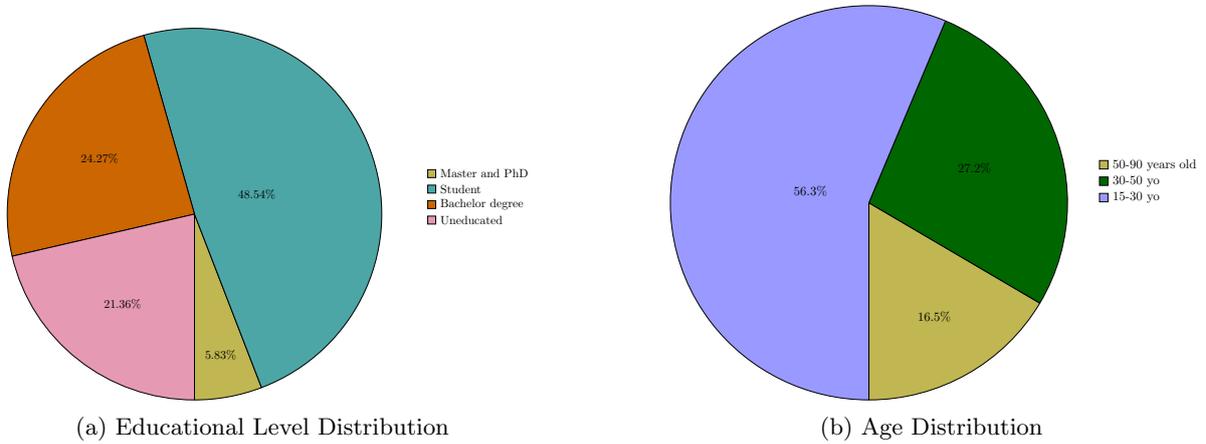

Figures \ref{fig:method_and_age_distribution} and \ref{fig:ageeducationdistribution} illustrate the gender of the speakers, the method of the interview, the education level of participants, and their age range, respectively.

\begin{figure}
	\centering
		\begin{subfigure}{0.45\textwidth}
			\centering
			\resizebox{\linewidth}{!}{%
				\begin{tikzpicture}
				\pie[radius=6, rotate=270, color={pink!70, purple!40!blue}, text=legend, every annotation/.style={font=\huge}]{51.46/Female, 48.54/Male}
				\end{tikzpicture}
			}
			\caption*{(a) Gender Distribution}
			\label{fig:gender_distribution}
		\end{subfigure}
		\hfill
		\begin{subfigure}{0.45\textwidth}
			\centering
			\resizebox{\linewidth}{!}{%
				\begin{tikzpicture}
				\pie[radius=6, rotate=360, text=pin, color={green!70!black, red!10}, text = legend,  every annotation/.style={font=\huge}]  
				{89.3/Face To Face,  10.7 /Online} 
				\end{tikzpicture}
			}
			\caption*{(b) Gathering Methods: Online vs. Face-to-Face}
			\label{fig:gathering_methods}
		\end{subfigure}
	\caption{The distribution of participants' genders and gathering method of the recording audios in Sorani Nas dataset}
	\label{fig:method_and_age_distribution}
\end{figure}
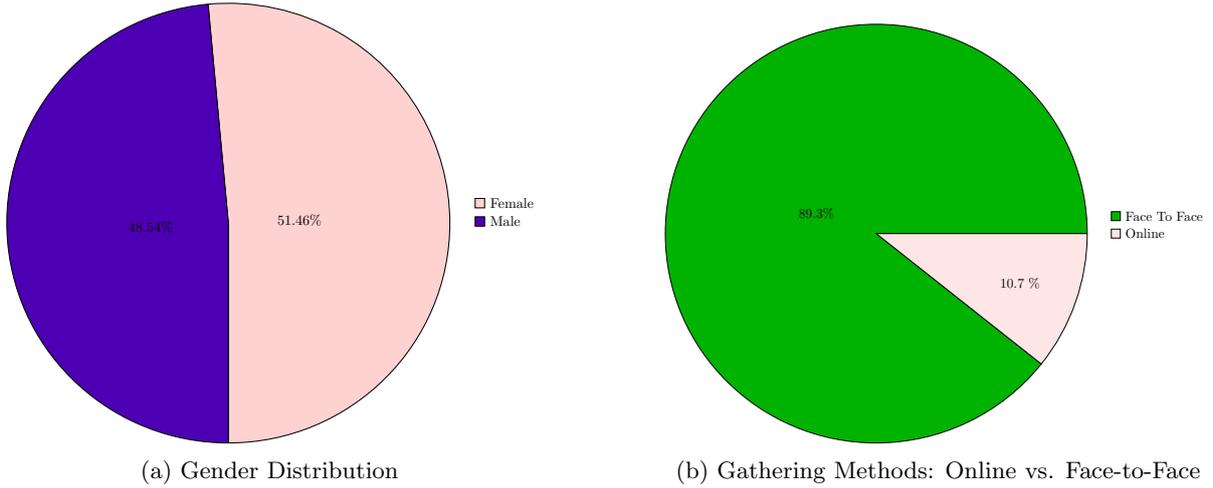

Table \ref{SpecificationsZarokay_Bawan_speech_dataset} provides an overview of the Sorani Nas dataset, briefly covering the important information discussed in previous sections.
\begin{table}
	\begin{center}
		\caption{Specifications of Sorani Nas speech dataset}
		
		\begin{tabular}{|p{4cm}|p{4cm}|}
			\hline
			{Title} & \fontsize{12}{12} {Value} 
			\\ \hline
			The dataset name & Sorani Nas(SN)                       
			\\ \hline
			Recording hardware & microphone           
			
			\\ \hline
			Recording software & Audacity         
			
			\\ \hline
			Duration & 29h 16m 40 sec         
			
			\\ \hline
			Number of speakers & 107          
			
			\\ \hline
			Average Duration of Speakers & 16.4m                              
			\\ \hline
			Sample rate & 44100 Hz                               
			\\ \hline
			Frequency & 22050 Hz                               
			\\ \hline
			Format & wav                         
			\\ \hline             
		\end{tabular}
		\label{SpecificationsZarokay_Bawan_speech_dataset}
	\end{center}
\end{table}

\begin{figure}
	\centering
	\fbox{  \includegraphics[width=14.5cm,height=6cm]{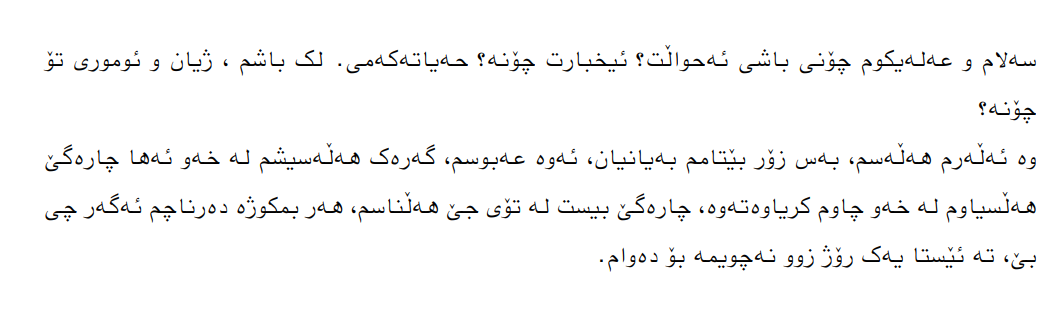}}
	\caption{A sample of transcribed Garmiani speech from Sorani Nas}
	\label{fig:garmiani_txt}
\end{figure}

\begin{figure}
	\centering
	\fbox{ \includegraphics[width=14.5cm,height=6cm]{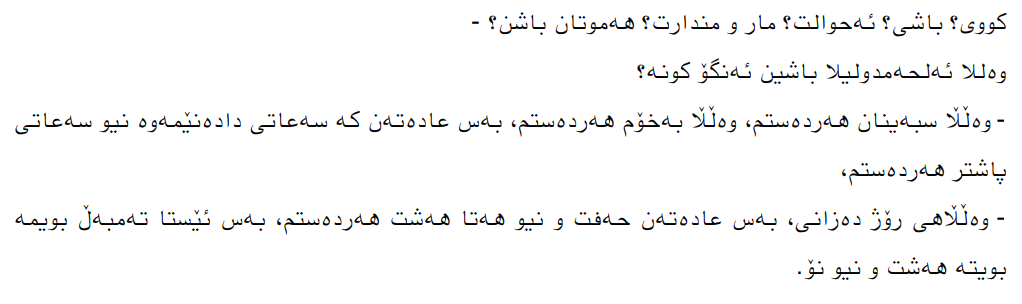}}
	\caption{A sample of transcribed Hewleri speech from Sorani Nas}
	\label{fig:hewleri_txt}
\end{figure}

\begin{figure}
	\centering
	\fbox{  \includegraphics[width=14.5cm,height=6cm]{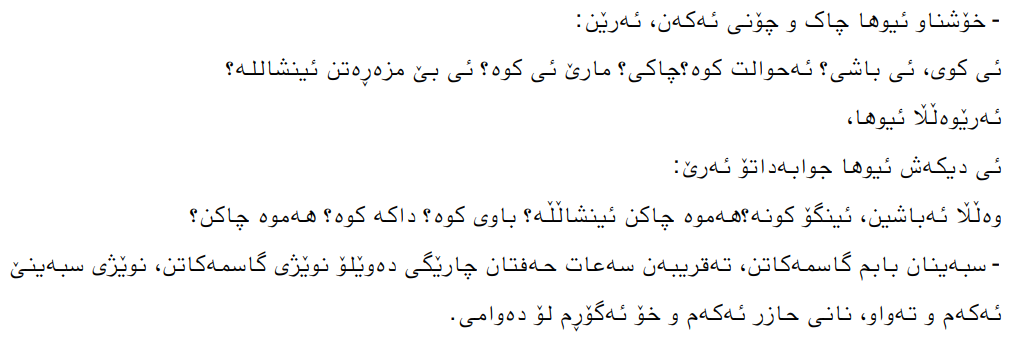}}
	\caption{A sample of transcribed Khoshnawi speech from Sorani Nas}
	\label{fig:khoshnawi_txt}
\end{figure}

\begin{figure}
	\centering
	\fbox{ \includegraphics[width=14.5cm,height=7cm]{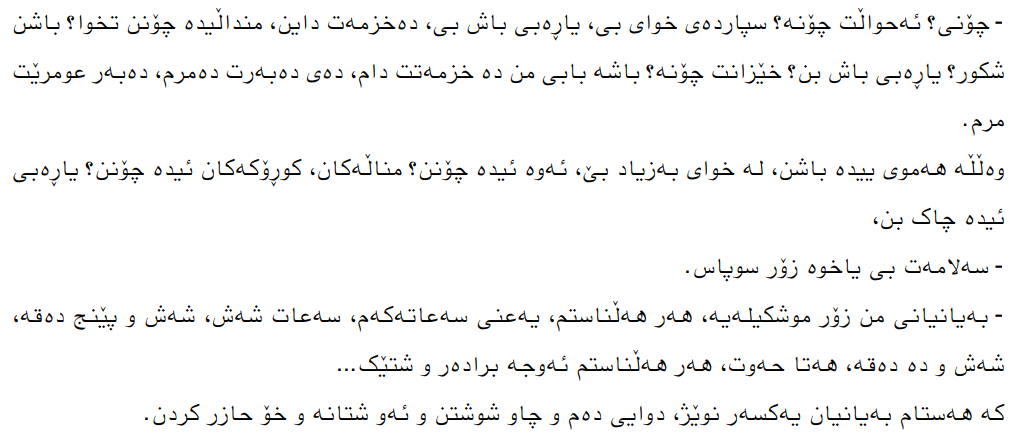}}
	\caption{A sample of transcribed Pishdari speech from Sorani Nas}
	\label{fig:pishdari_txt}
\end{figure}

\begin{figure}
	\centering
	\fbox{  \includegraphics[width=14.5cm,height=6cm]{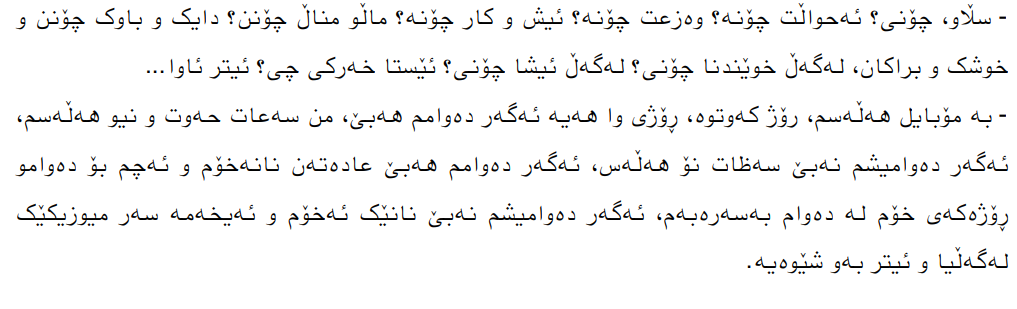}}
	\caption{A sample of transcribed Sulaimani speech from Sorani Nas}
	\label{fig:sulaimani_txt}
\end{figure}

\begin{figure}
	\centering
	\fbox{\includegraphics[width=14.5cm,height=6cm]{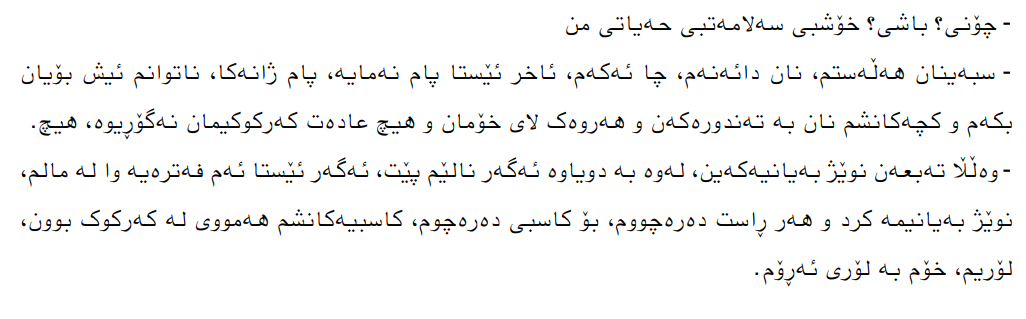}}
	\caption{A sample of transcribed Karkuki speech from Sorani Nas}
	\label{fig:karkuki_txt}
\end{figure}

Figure \ref{subdialects_dostribution} provides a summary of the subdialect percentages in our data set Sorani Nas dataset, and Figures~\ref{fig:garmiani_txt}, \ref{fig:hewleri_txt}, \ref{fig:khoshnawi_txt}, \ref{fig:pishdari_txt}, \ref{fig:sulaimani_txt}, and \ref{fig:karkuki_txt} show prescribed samples of the recordings, which are about greetings and their morning routines, for the studied subdialects.



\subsection{Balancing The Dataset}
The generated audio samples ranged in length from six minutes to forty-five minutes. To generate a dataset with wider diversity, we segmented the recordings into discrete time intervals, namely 1-second, 3-second, 5-second, 10-second, and 30-second segments. 

\begin{figure}
	\centering
	\fbox{%
		\begin{subfigure}[b]{0.45\textwidth}
			\includegraphics[width=\textwidth]{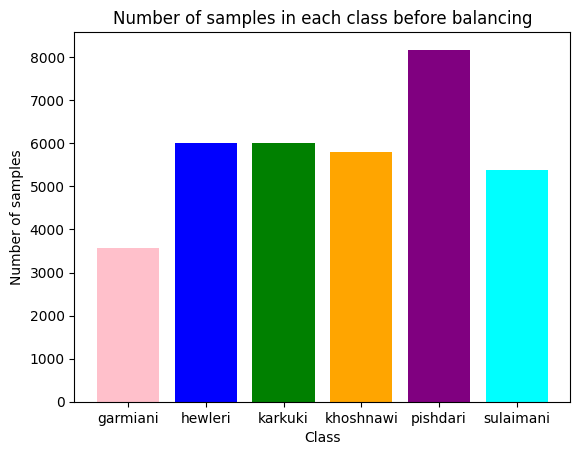}
			\caption*{(a) Imbalanced Sorani Nas}
			\label{fig:image1}
		\end{subfigure}%
		\hspace{0.08\textwidth}
		\begin{subfigure}[b]{0.45\textwidth}
			\includegraphics[width=\textwidth]{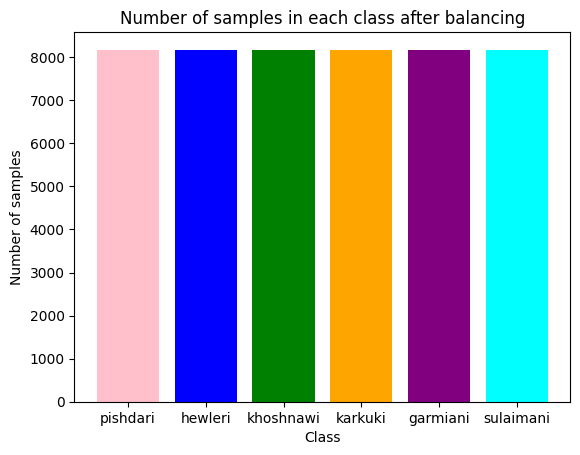}
			\caption*{(b) Balanced Sorani Nas}
			\label{fig:image2}
		\end{subfigure}%
	}
	\caption{Handling class imbalance in the Sorani Nas dataset using Random Oversampling}
	\label{oversampling zarybawan}
\end{figure}

We attempted to balance Sorani Nas with two techniques: undersampling and oversampling. The oversampling approach, as presented in Figure \ref{oversampling zarybawan}, depicts the distribution of classes within the imbalanced Sorani Nas dataset, particularly emphasising the subdialect classes. The dataset displays a notable imbalance among the subdialect classes, with dissimilar quantities of samples for each class.
The Random Oversampling technique was implemented on the dataset's 3-second duration samples to minimise this issue. Consequently, the class distribution was altered, increasing the number of samples for each subdialect class to 8172, which, in this case, is the Pishdari subdialect, representing the maximum number of samples across all classes. The process of equalising sample counts was undertaken to mitigate the class imbalance and ensure a more equitable portrayal of the subdialects present in the dataset.

\par \begin{figure}
	\centering
	\fbox{%
		\begin{subfigure}[b]{0.45\textwidth}
			\includegraphics[width=\textwidth]{3im.png}
			\caption*{(a) Imbalanced Sorani Nas}
			\label{fig:imagee1}
		\end{subfigure}%
		\hspace{0.08\textwidth}
		\begin{subfigure}[b]{0.45\textwidth}
			\includegraphics[width=\textwidth]{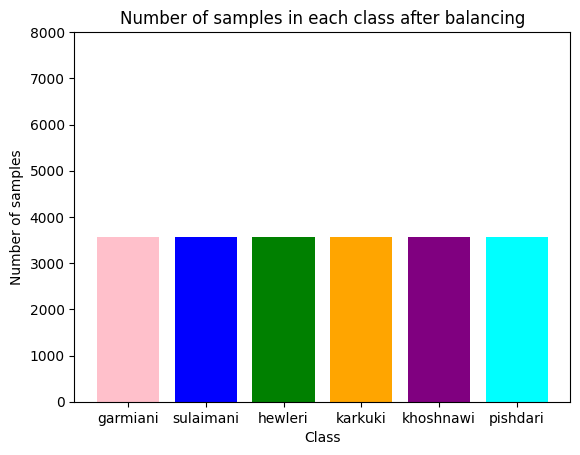}
			\caption*{(b) Balanced Sorani Nas}
			\label{fig:imagee2}
		\end{subfigure}%
	}
	\caption{Handling Class Imbalance in the Sorani Nas dataset using Undersampling}
	\label{undersampling zarybawan}
\end{figure}

Figure \ref{undersampling zarybawan} depicts the initial imbalanced distribution of subdialect classes in the Sorani Nas dataset. The dataset displayed heterogeneous sample sizes across the subdialect classes before applying undersampling. The observed difference was notably conspicuous, as the Garmiani subdialect had the smallest sample size, amounting to only 3566 samples. To tackle this problem, the dataset's 3-second duration samples were subjected, as an example, to an undersampling technique. The undersampling process included randomly selecting a subset of samples from the majority class, reducing its quantity to align with the number of samples in the minority class. Consequently, the dataset was adjusted to ensure equal representation of each subdialect class by standardising the number of samples to 3566. This approach aimed to achieve a more equitable distribution of subdialects in the dataset.

\begin{figure}
	\centering
	\includegraphics[width=11cm]{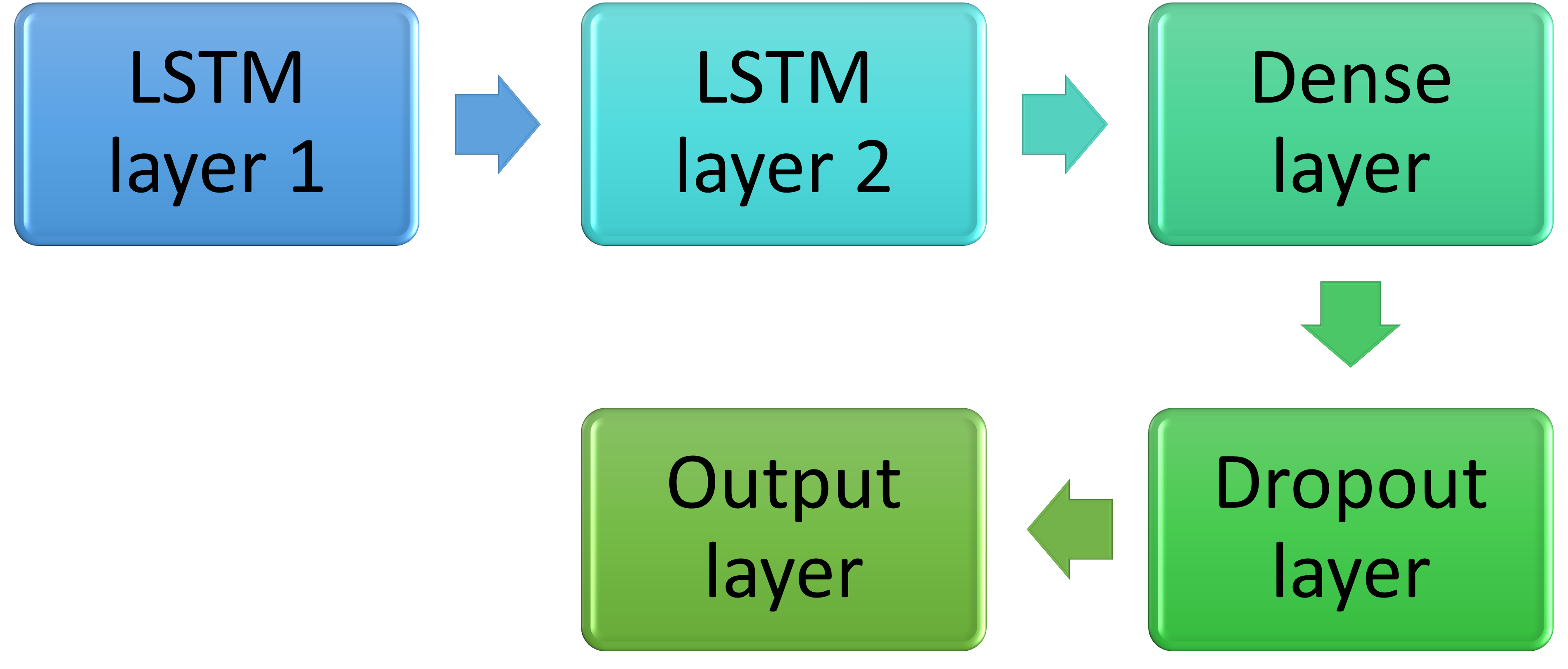}
	\caption{RNN-LSTM full steps}
	\label{fig:RNNLSTMdesign}
\end{figure}

Six distinct datasets were generated by segmenting the audio into durations of 1 second, 3 seconds, 5 seconds, 10 seconds and 30 seconds. To guarantee optimal clarity and quality, the audio files are saved in the \textit{wav} file format, utilising a bit rate of 1411 kbps. 

\subsection{Experiments Using ANN}
\begin{figure}
	\centering
	\begin{subfigure}{0.85\textwidth}
		\centering
		\includegraphics[width=\linewidth, height=5cm]{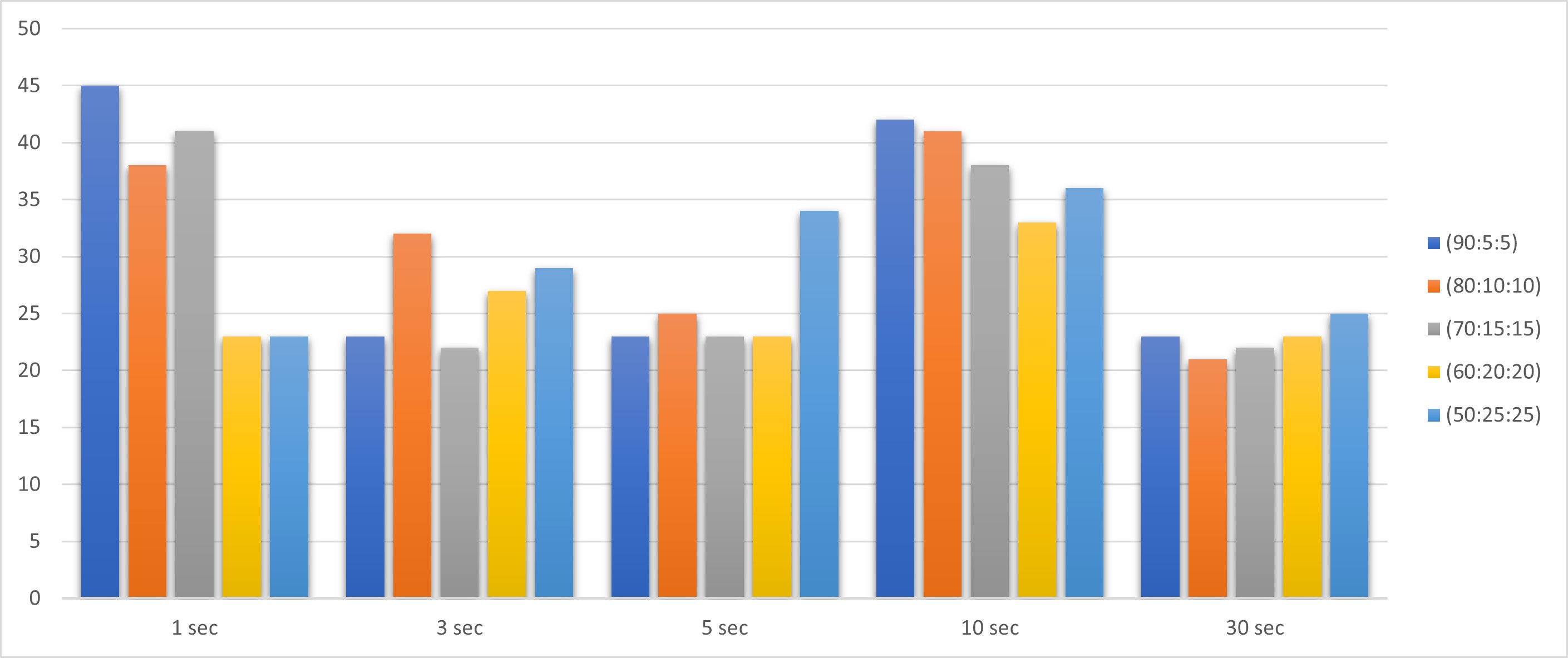}
		\caption*{(a) Imbalanced dataset with the different track durations and different dataset splitting}
	\end{subfigure}
	\hfill
	\begin{subfigure}{0.85\textwidth}
		\centering
		\includegraphics[width=\linewidth, height=5cm]{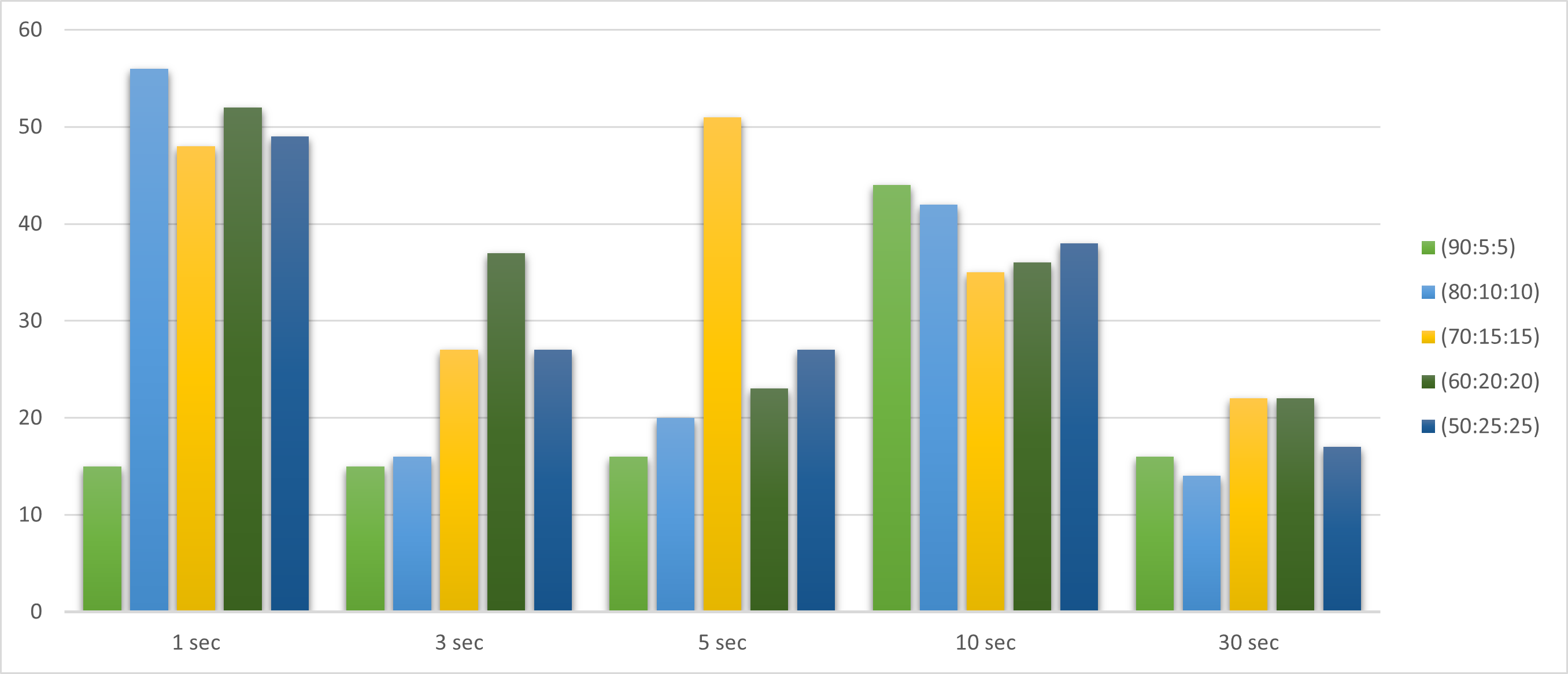}
		\caption*{(b) Undersampled dataset with the different track durations and different dataset splitting}
	\end{subfigure}
	
	\begin{subfigure}{0.85\textwidth}
		\centering
		\includegraphics[width=\linewidth, height=5cm]{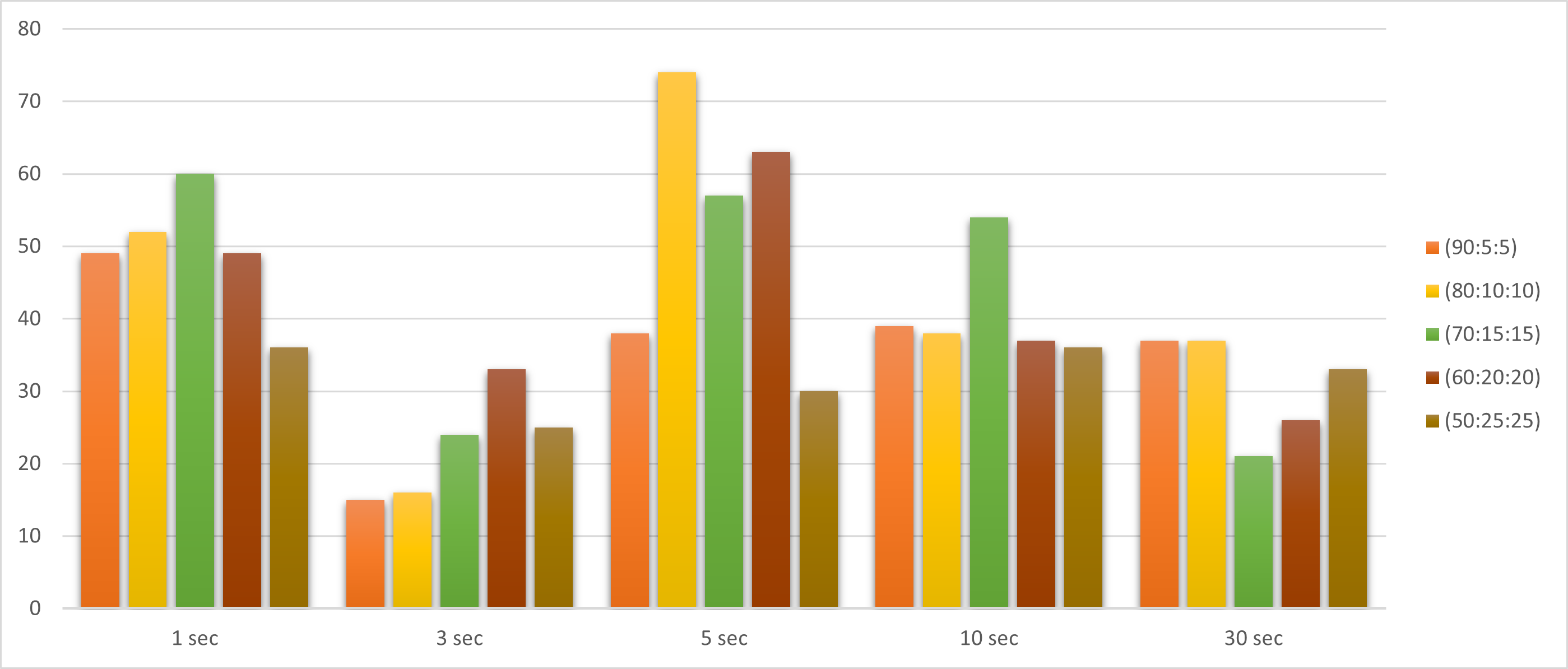}
		\caption*{(c) Oversampled dataset with the different track durations and different dataset splitting}
	\end{subfigure}
	
	\caption{ANN model accuracies}
	\label{fig:ANN_Accuracies}
\end{figure}

We conducted a comprehensive set of experiments on an ANN model, exploring various combinations of dataset durations and training/testing set distributions. Specifically, we evaluated five different dataset track durations, including 1-second, 3-second, 5-second, 10-second and 30-second segments. We make a version of the dataset on each one. Additionally, we examined each one of them on different ratios of dataset splitting into training and testing sets, namely 90:10, 80:20, 70:30, 60:40 and 50:50. Furthermore, we investigated three different dataset types: an imbalanced dataset, a balanced dataset with an oversampling technique and a balanced dataset with an undersampling technique. We conducted 75 experiments on the ANN model, which are shown in Figure \ref{fig:ANN_Accuracies}, each with the predefined training model parameters described in Table \ref{model_Parameters}.

Among the experiments conducted on the imbalanced dataset, After conducting our experiments, we found that the highest accuracy was achieved when using the oversampled dataset with 5-second audio segments and an 80:10:10 dataset splitting ratio. This configuration resulted in an accuracy of 56\%. Additionally, when using the undersampled dataset with 1-second segments and an 80:20 dataset splitting ratio, we achieved an accuracy of 45\%. Similarly, when using 1-second segments with a 90:10 dataset splitting ratio on the imbalanced dataset, the accuracy was 45\%.

On the other hand, the lowest accuracy was observed with the undersampled Sorani Nas dataset, using 1-second segments and a 90:10 training and testing set ratio, resulting in an 

accuracy of 15\%. The same low accuracy of 23\% was obtained under similar circumstances but with the imbalanced dataset. Furthermore, an accuracy of 15\% was obtained when using 3-second segments with 90:10 and 80:20 dataset splitting ratios on the oversampled Sorani Nas dataset.
Considering all the experiments conducted, the accuracies were generally lower for longer dataset durations, especially with 30-second segments. Similarly, accuracy tended to be lower when dataset splitting approaches 50:50 and 60:40. These findings suggest that shorter segment durations, combined with an overbalanced dataset and a higher proportion of training samples, yield better accuracy rates for the ANN model. 

\subsection{Experiments Using CNN}
\begin{figure}
	\centering
	\fbox{ \includegraphics[width=14cm, height=10cm]{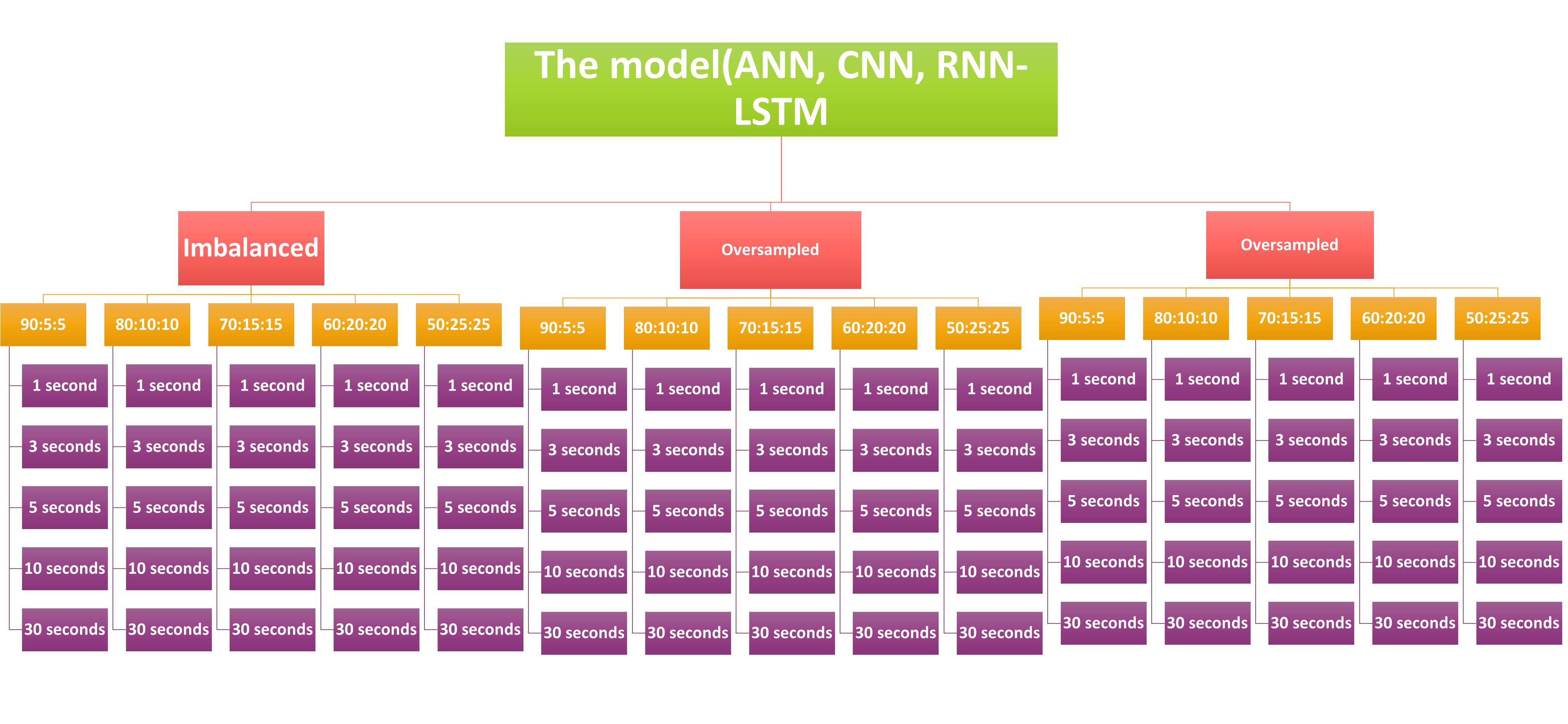}}
	\caption{The experiments done on each model}
	\label{fig:experiements}
\end{figure}

\begin{figure}
	\centering
	\begin{subfigure}{0.85\textwidth}
		\centering
		\includegraphics[width=\linewidth, height=5cm]{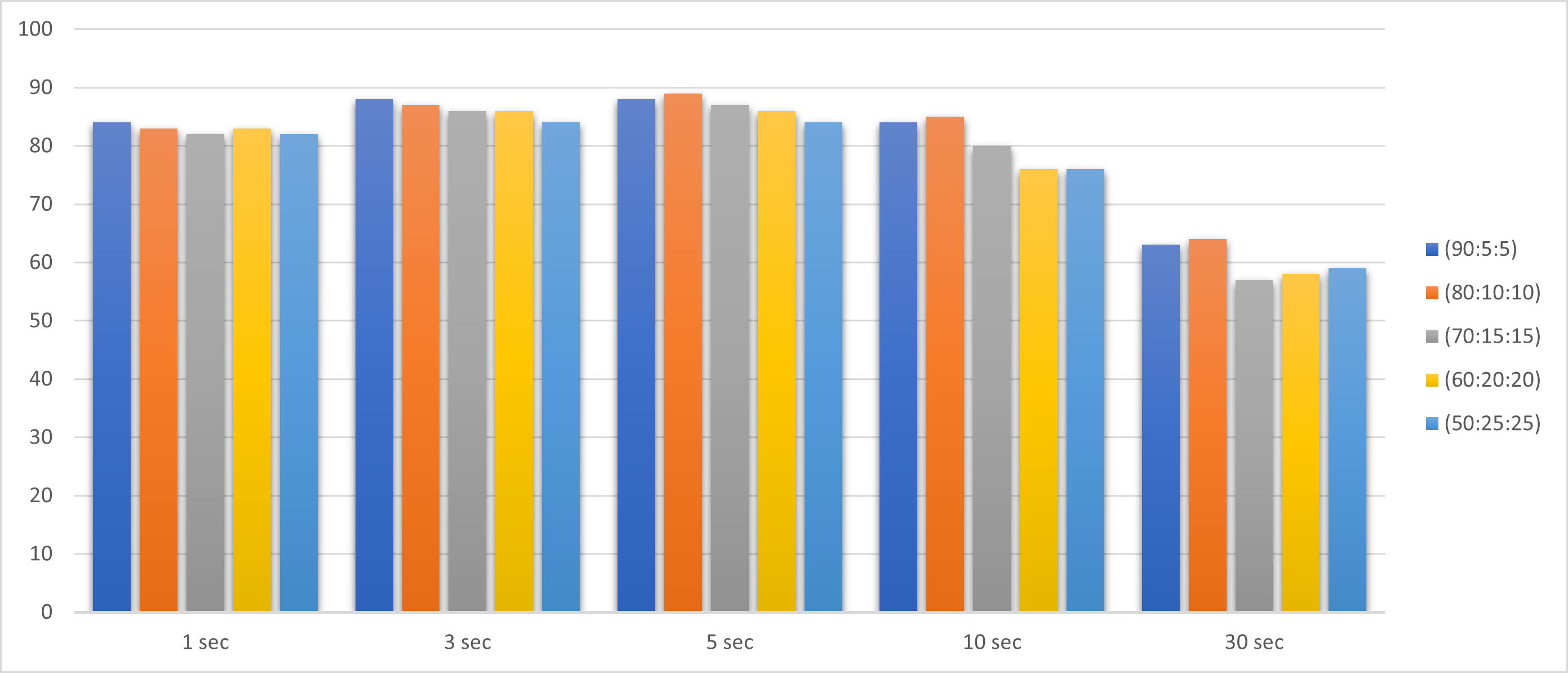}
		\caption* {(a) Imbalanced dataset with the different track durations and different dataset splitting}
	\end{subfigure}
	\hfill
	\begin{subfigure}{0.85\textwidth}
		\centering
		\includegraphics[width=\linewidth, height=5cm]{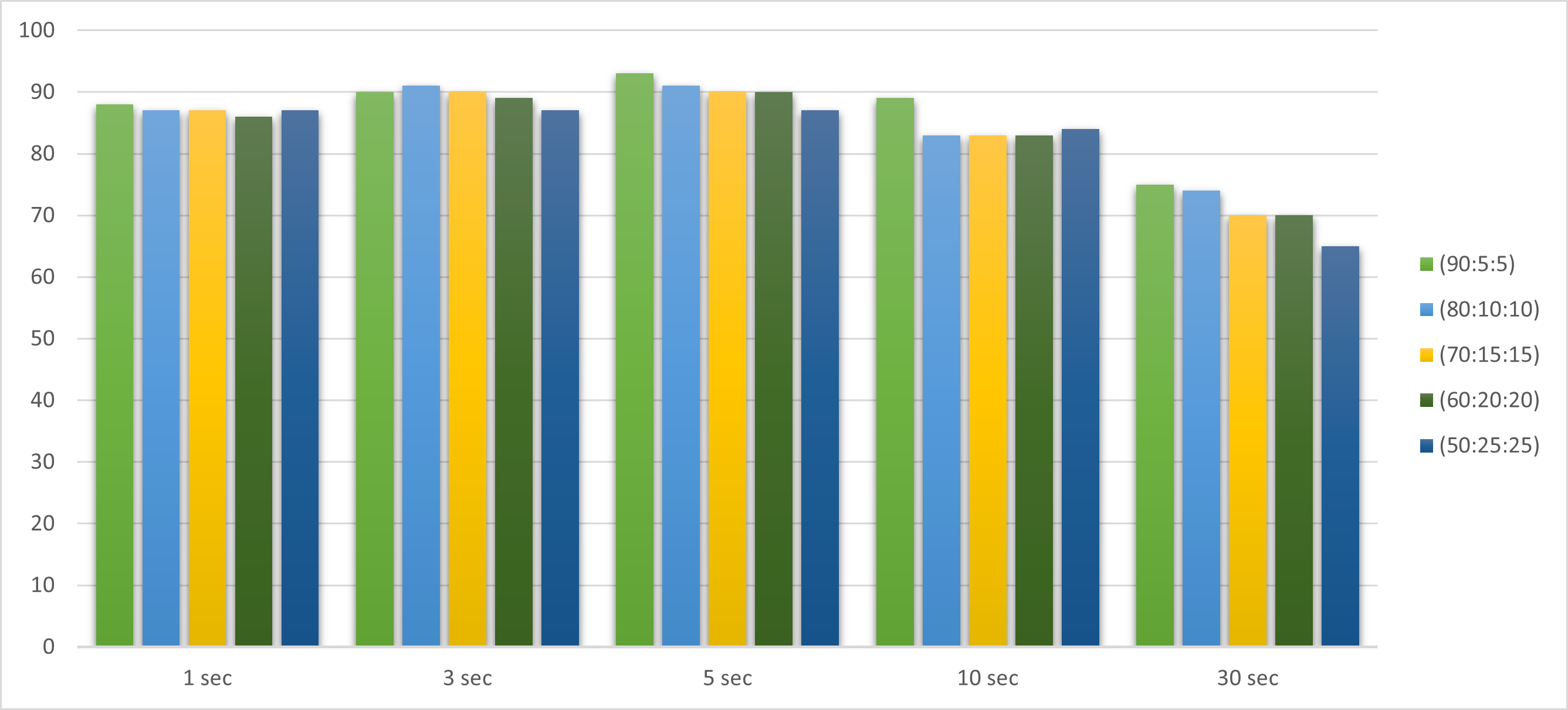}
		\caption*{(b) Undersampled dataset with the different track durations and different dataset splitting}
	\end{subfigure}
	
	\begin{subfigure}{0.85\textwidth}
		\centering
		\includegraphics[width=\linewidth, height=5cm]{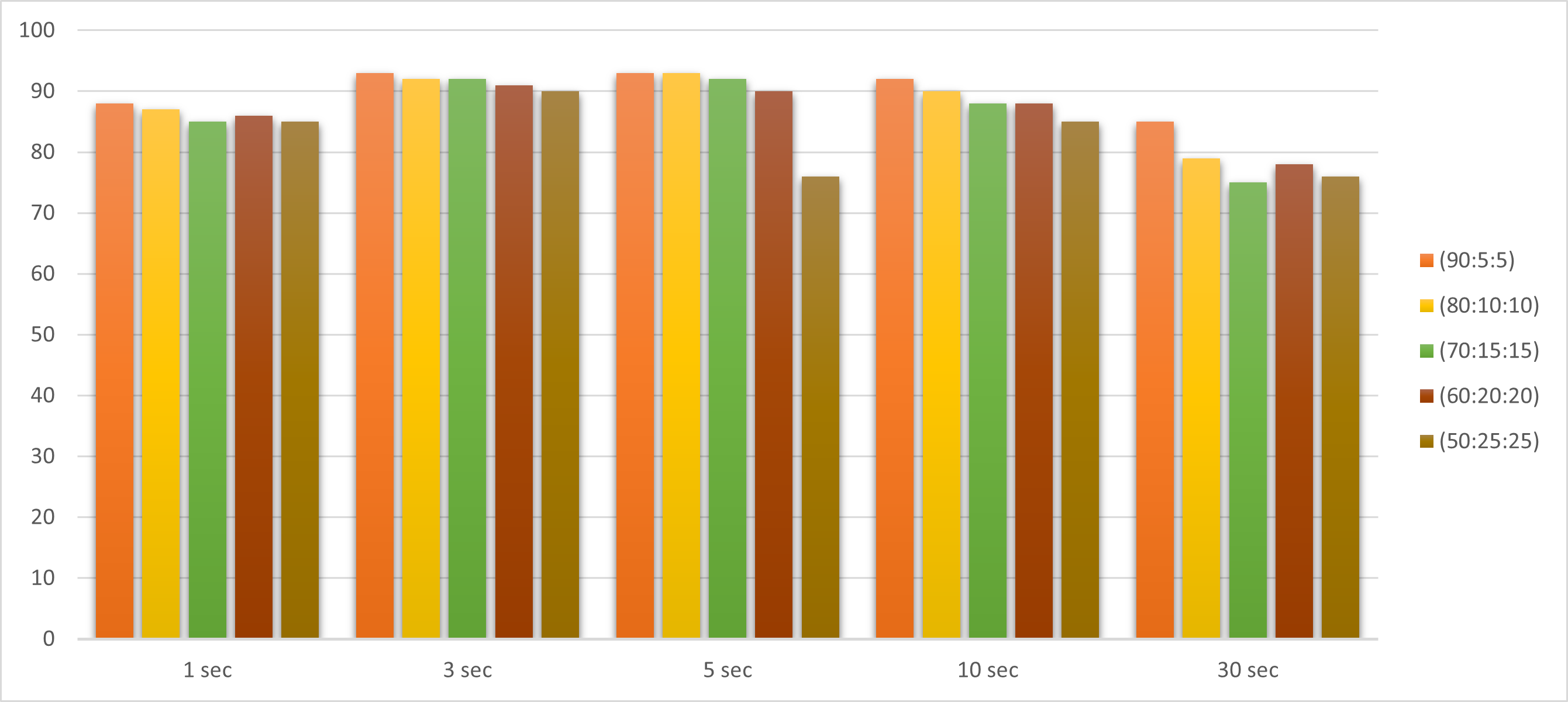}
		\caption*{(c) Oversampled dataset with the different track durations and different dataset splitting}
	\end{subfigure}
	
	\caption{CNN model accuracies}
	\label{fig:CNN_Accuracies}
\end{figure}

The CNN model underwent an experimentation process similar to that of the ANN model, including various durations of the segments (1 second, 3 seconds, 5 seconds, 10 seconds and 30 seconds). Additionally, each version of the audio segment duration dataset was tested using five different distributions for the training, validation and testing sets: 90:5:5, 80:10:10, 70:15:15, 60:20:20 and 50:25:25. All experiments with three different types of datasets were employed: imbalanced dataset, balanced with oversampling technique and balanced with undersampling technique.\\ We conducted 75 experiments on the CNN model, as represented in Figure \ref{fig:experiements}, and their accuracies are found in Figure \ref{fig:CNN_Accuracies}, each with the predefined training model parameters described in Table \ref{model_Parameters}. The oversampled dataset, consisting of 3 and 5-second segments, achieved the highest accuracy of 93\%. This was observed in the dataset splitting ratios of 90:5:5. Close behind, an accuracy of 92\% was attained using 3-second and 5-second segments in the dataset splitting’s of 80:10:10 and 70:15:15. For the undersampled dataset, the highest accuracy of 93\% was achieved with 5-second sound durations in both the 80:10:10 and 90:5:5 dataset splitting ratios. On the other hand, the best accuracy for the imbalanced dataset was 89\% in the 80:10:10 dataset splitting with 5-second segments.

Conversely, the lowest accuracy observed was 57\% when utilising the imbalanced dataset with 30-second segments, particularly in the 70:15:15 distribution. For the unbalanced dataset, accuracy ranged from 65\% to 75\% with 30-second segments across various dataset-splitting versions. The worst case for oversampled datasets was 75\% to 78\% precision with 30-second track durations in almost all data set splitting ratios except 90:5:5.

Furthermore, our overall observations support the earlier notion that the imbalanced dataset tends to yield the lowest accuracy rates while balancing techniques such as undersampling and oversampling result in increased accuracy \cite{hernandez2013empirical}. In particular, the CNN model demonstrated superior performance to the ANN model.

\begin{figure}
	\centering
	\begin{subfigure}[t]{0.49\textwidth}
		\centering
		\includegraphics[width=\linewidth, height=5cm]{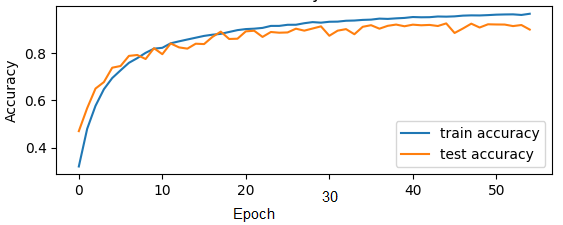}
		\caption* {(a) Error/Loss analysis during CNN training}
	\end{subfigure}
	\hfill
	\begin{subfigure}[t]{0.49\textwidth}
		\centering
		\includegraphics[width=\linewidth, height=5cm]{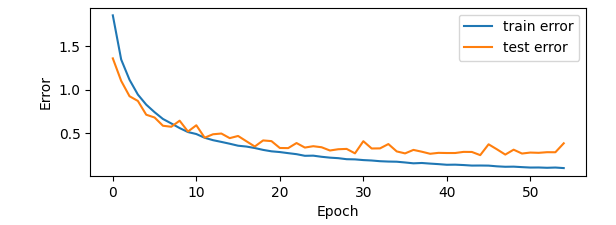}
		\caption* {(b) Training and testing accuracy analysis during CNN training}
	\end{subfigure}

	\caption{Error/Loss and training, testing analysis during CNN Training}
	\label{fig:cnn_err}
\end{figure}

In addition, Figure \ref{fig:cnn_err} shows how training and testing CNN model errors gradually decrease with increasing epochs in the CNN model. As the model is trained on the training data, it tries to minimise the error or loss function, resulting in lower errors over time. The decreasing trend of training and testing errors indicates that the model is learning and improving its performance. Finally, implementing Earlystopping technique causes the training process to halt when no improvement was observed for 10 consecutive epochs.
\begin{table}
	\begin{center}
		\caption{Model parameters}
		
		\begin{tabular}{|p{4cm}|p{4cm}|}
			\hline
			\fontsize{12}{12} { Parameter} & \fontsize{12}{12} { Value }
			\\ \hline
			Learning\_rate & 0.0001 \\
			\hline
			Batch\_size & 32 \\
			\hline
			Epochs & 200 \\
			\hline
			Patience & 10 
			\\ \hline             
		\end{tabular}
		\label{model_Parameters}
	\end{center}
\end{table}

\subsection{Experiments Using  RNN-LSTM}
In our experiment with the RNN-LSTM model, we carried out the same 75 experiments as we did on the CNN model. All are illustrated in Figure \ref{fig:experiements}, and their accuracies are represented in Figure \ref{fig:CNN_Accuracies}. We achieved a remarkable accuracy of 96\% by employing an oversampled dataset, an 80:10:10 dataset splitting ratio, and utilising 5-second track segments. Similarly, a high accuracy of 95\% was obtained when using 3-second track segments, an 80:10:10 distribution, a 5-second segment duration, and a 90:10:10 dataset splitting ratio. For the undersampled dataset, the highest accuracy observed was 93\% with 5-second sound durations in both the 80:10:10 and 90:5:5 dataset splitting ratios. On the other hand, the best accuracy for the imbalanced dataset was 92\% with an 80:10:10 dataset splitting ratio and 5-second segments.

\begin{figure}
	\centering
	\begin{subfigure}[t]{0.49\textwidth}
		\centering
		\includegraphics[width=\linewidth, height=4cm]{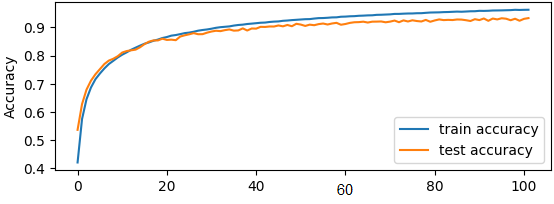}
		\caption*{(a) Error/Loss analysis}
	\end{subfigure}
	\hfill
	\begin{subfigure}[t]{0.49\textwidth}
		\centering
		\includegraphics[width=\linewidth, height=4cm]{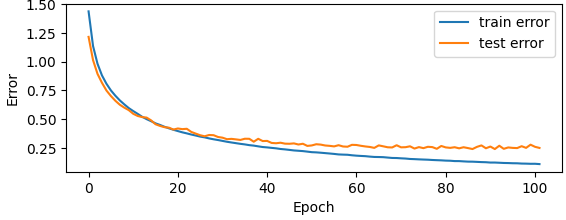}
		\caption*{(b) Training and testing accuracy }
	\end{subfigure}  
	\caption{Error/Loss and training, testing analysis during RNN-LSTM training}
	\label{fig:rnn_err}
\end{figure}

Conversely, the lowest accuracy observed was 51\%, which occurred when utilising the imbalanced dataset with 30-second segments, particularly in the 80:10:10 dataset splitting ratio. For the unbalanced dataset, an accuracy of 55\% was achieved with 30-second segments in the 60:20:20 dataset splitting ratio. Regarding the overbalanced datasets, the worst-case accuracy ranged from 75\% to 78\% with 30-second track durations in almost all dataset-splitting ratios except for the 90:5:5 ratio. Moreover, the gradual decrease of training and testing errors  with increasing epochs in the RNN-LSTM model is illustrated in Figure \ref{fig:rnn_err}. The model minimises the error or loss function and increases the training/testing accuracies during training by learning from the training data, gradually reducing errors and enhancing its efficacy. The training process terminates when no improvement is observed for 10 consecutive epochs due to utilising early stopping technique.

\begin{figure}
	\centering
	\begin{subfigure}{0.85\textwidth}
		\centering
		\includegraphics[width=\linewidth, height=5cm]{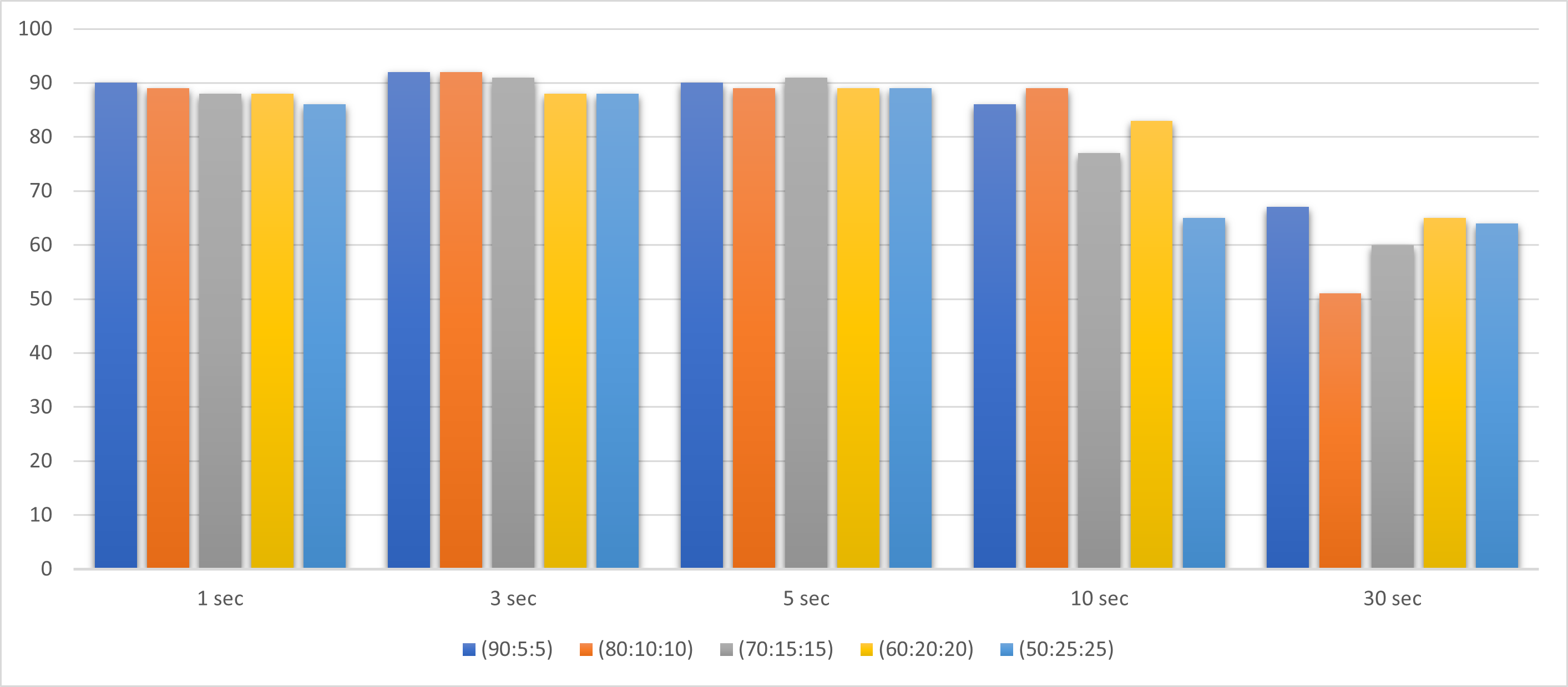}
		\caption*{(a) Imbalanced dataset with the different track durations and different dataset splitting}
	\end{subfigure}
	\hfill
	\begin{subfigure}{0.85\textwidth}
		\centering
		\includegraphics[width=\linewidth, height=5cm]{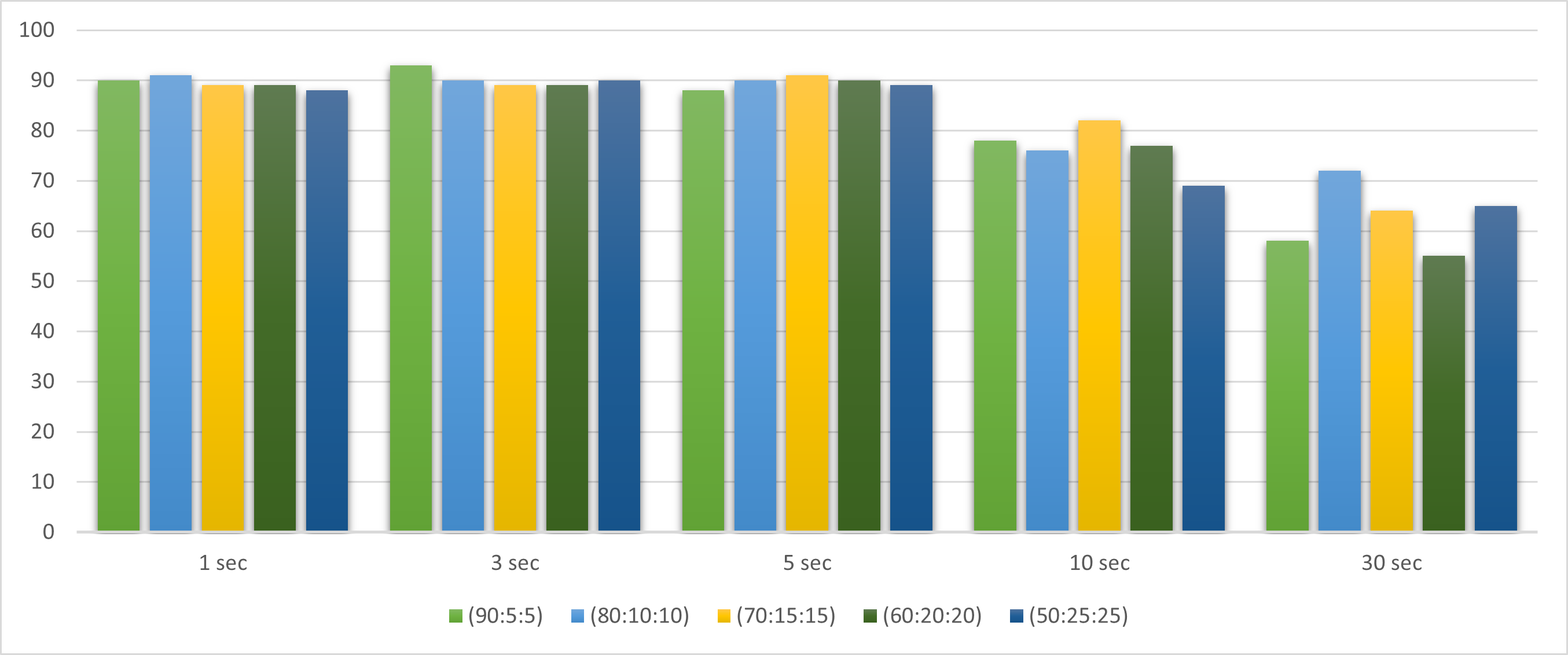}
		\caption*{(b)undersampled dataset with the different track durations and different dataset splitting}
	\end{subfigure}
	
	\begin{subfigure}{0.85\textwidth}
		\centering
		\includegraphics[width=\linewidth, height=5cm]{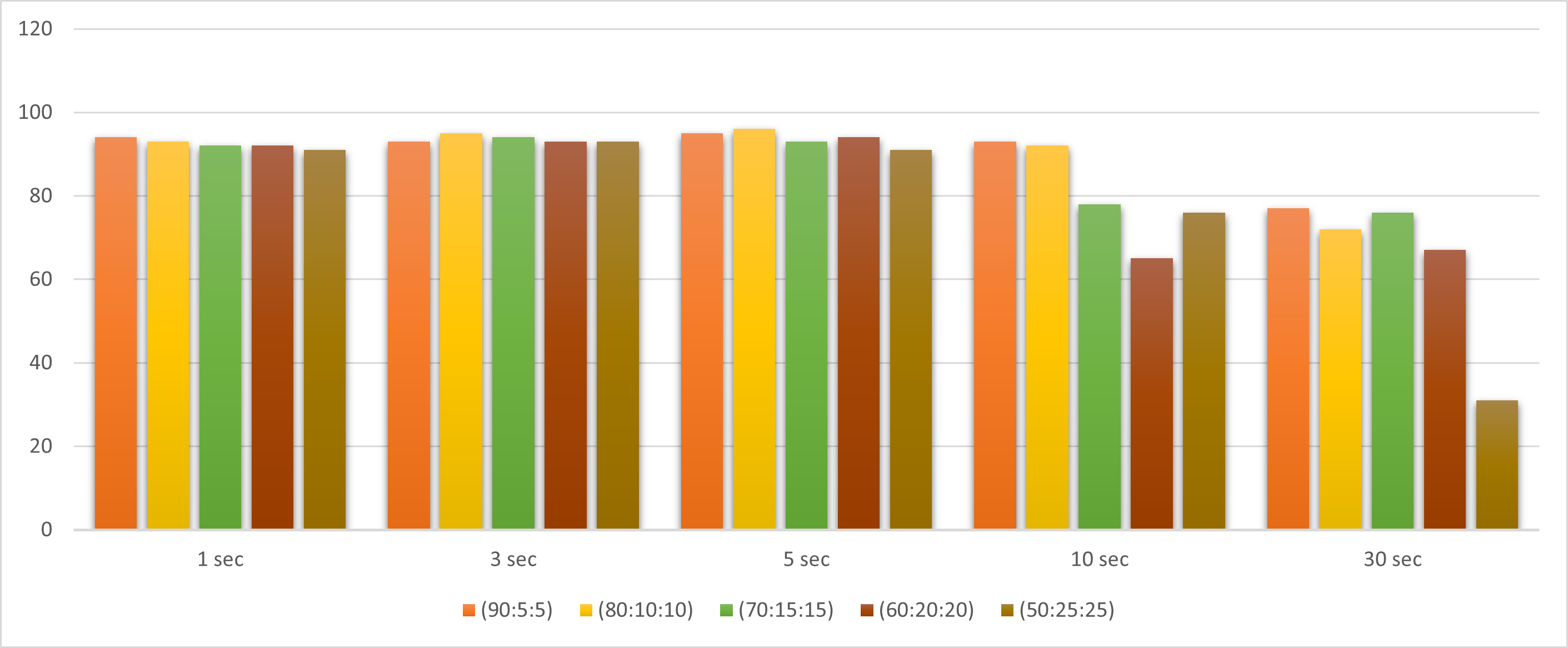}
		\caption*{(c) Oversampled dataset with the different track durations and different dataset splitting}
	\end{subfigure}
	
	\caption{RNN-LSTM model accuracies}
	\label{fig:RNN_LSTM_Accuracies}
\end{figure}
Ultimately, we observed that the RNN-LSTM model consistently outperformed the CNN model in accuracy.

\subsection{Evaluation and Discussion}
We encountered several challenges during the classification and dataset development of Kurdish-Sorani subdialects. One major difficulty was the lack of a comprehensive automated classification system for subdialects within a specific dialect, which made it challenging to compare our results with existing studies. There has been limited research on dialects in general, and especially on the subdialects between Hewleri and Sulaimani. Furthermore, our interaction with individuals posed another challenge. Despite assuring them that the recordings would be used solely for research purposes and would not be shared on social media, some participants were hesitant to have their voices recorded. The winter season also added to the difficulties, as it was challenging to visit cities and villages, particularly those located in mountainous regions, due to poor road conditions and frequent cloud cover. These factors made it difficult to gather data effectively. Moreover, conducting visits and returning on the same day proved problematic as it was often necessary to make multiple trips to collect sufficient speech recordings. Despite these challenges, we developed the dataset and conducted a total of 225 experimental observations.

\begin{figure}
	\centering
	\includegraphics[width=14.5cm, height=7.5cm]{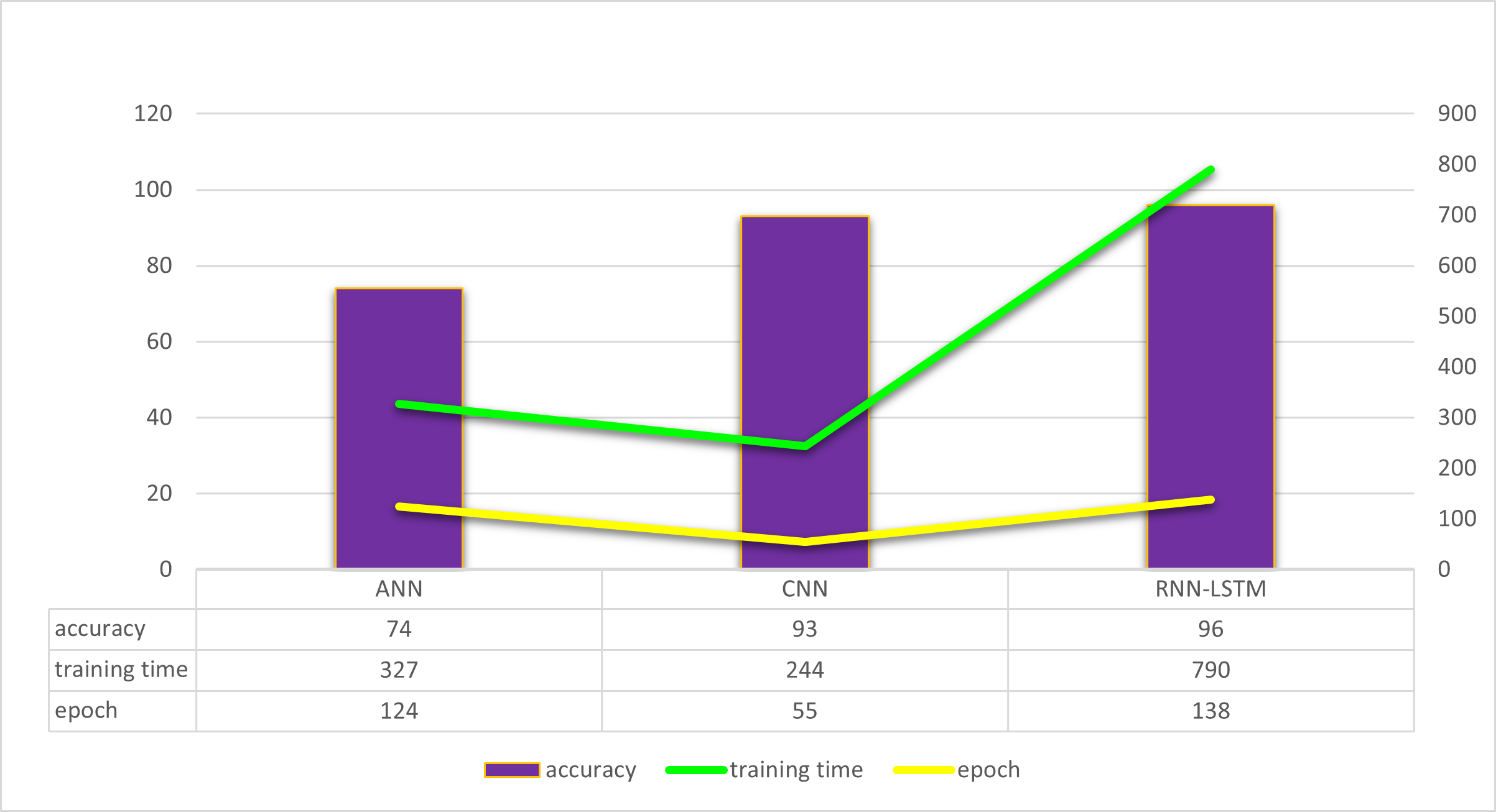}
	\caption{Comparing ANN, CNN, and RNN-LSTM models}
	\label{fig:compare_models}
\end{figure}

Based on the 225 experimental observations we conducted experiments with three different models, 75 on each model, as represented in Figure \ref{fig:experiements}. Imbalanced, undersampled and oversampled datasets. Each one of them with 5 different dataset splitting's into training, validation and testing sets and three models on each one of them, namely ANN, CNN and RNN-LSTM on subdialect classification using Sorani Nas, 75 trails on each model, because we tried three different datasets, imbalanced, oversampled and undersampled datasets, As shown in Figure \ref{fig:compare_models} it was observed that the highest accuracy rates were get when utilising an oversampled dataset. Each segment of the dataset had a duration of 5 seconds. The dataset was split into training, validation and testing sets using a ratio of 80:10:10 for RNN-LSTM and CNN and a ratio of 80:20 for ANN.

In addition, mostly the splitting’s that gives good accuracy are 90:5:5, 80:10:10 and 70:15:15 dataset splittings, with durations of 1 second, 3 seconds and 5 seconds, which repeated the experiment result in \newcite{gultekin13turkish}  which  found out that 3 seconds perform much better than shorter ones, on the contrary, The experiments that give low accuracies are 30 seconds mostly,with the 50:25:25 and 60:20:20 dataset splitting as can be observed in Figure \ref{fig:ANN_Accuracies}, Figure \ref{fig:CNN_Accuracies} and Figure \ref{fig:RNN_LSTM_Accuracies}.

Furthermore, the accuracy of RNN-LSTM achieved the highest in all the situations \cite{sunny2020deep}, and RNN’s have the benefit of training faster and using less computational resources. That’s because there are fewer tensor operations to compute \cite{bansal2018low}. The training time in seconds of neural network models, such as RNN-LSTM, CNN and ANN, can vary depending on factors such as model complexity and architecture. Through experimentation, it has been observed that RNN-LSTM generally requires more training time compared to CNN, while CNN exhibits the minimum training time among the proposed models. This indicates that the computational requirements and the number of parameters in RNN-LSTM contribute to its longer training duration. Still, all the models, in general, don't need so much time because of using MFCC feature extractor \cite{ou2004text}. In contrast, the specialised structure of CNN allows for more efficient training. The respective training times for each model are presented in Figure \ref{fig:compare_models}.

Moreover, employing the early stopping technique in all three models ensures that training will cease if there is no improvement in accuracy for consecutive 10 epochs. Consequently, the specific epoch number at which each model stops training will differ based on various factors that mentioned earlier.

Table \ref{Classification metrics} displays the performance metrics of a classification model for different classes. Precision measures the accuracy of positive predictions, recall evaluates the ability to identify positive instances correctly, F1-score is a measure of a model's accuracy in binary classification tasks. It's the harmonic mean of precision and recall, two other important metrics in classification \cite{chicco2020advantages}, which provides a balanced assessment considering false positives, false negative,  true positives and true negatives. Each row represents a class, with precision, recall and F1-Score values specified. These metrics help evaluate the model's effectiveness in accurately classifying instances into different classes, with higher values indicating better performance.
\begin{table}
	\begin{center}
		\caption{Classification metrics for each subdialect in RNN-LSTM model}
		
		\begin{tabular}{|p{4.4cm}|p{3.4cm}|}
			\hline
			{\fontsize{12}{12} {Class}} &  {\fontsize{12}{12} {F1-Score}}
			\\ \hline
			Garmiani &  0.98 \\ \hline
			Hewleri &  0.94 \\ \hline
			Karkuki & 0.94 \\ \hline
			Khoshnawi & 0.93 \\ \hline
			Pishdari  & 0.91 \\ \hline
			Sulaimani&  0.92 \\ \hline
			
		\end{tabular}
		\label{Classification metrics}
	\end{center}
\end{table}
Finally,  we conducted validation on a subset of the dataset using two different approaches: human validation and machine prediction. Due to limited resources explaining the differences between Sorani subdialects, our validation process relied solely on human judgment. We played audio samples to native speakers of the Sorani subdialects and asked them to identify the corresponding subdialect. In some cases, they provided two subdialect options as they couldn't determine an individual one. On the other hand, for machine prediction, we used our trained model to predict the subdialect of individual audio samples. Table \ref{Detecting single tracks from each subdialect result} provides an example of the results obtained from both approaches for each subdialect.

\begin{figure}
	\centering
	\begin{minipage}{0.48\textwidth}
		\centering
		\includegraphics[width=\linewidth]{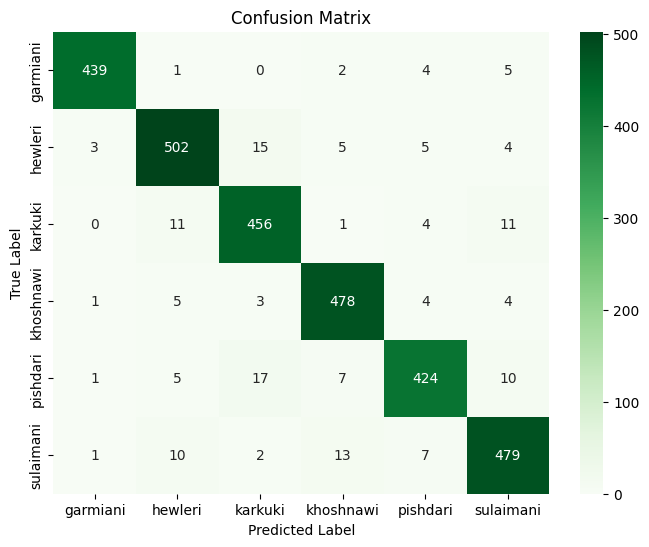}
		\caption*{(a) Confusion matrix of ANN model}
	\end{minipage}
	\hfill
	\begin{minipage}{0.48\textwidth}
		\centering
		\includegraphics[width=\linewidth]{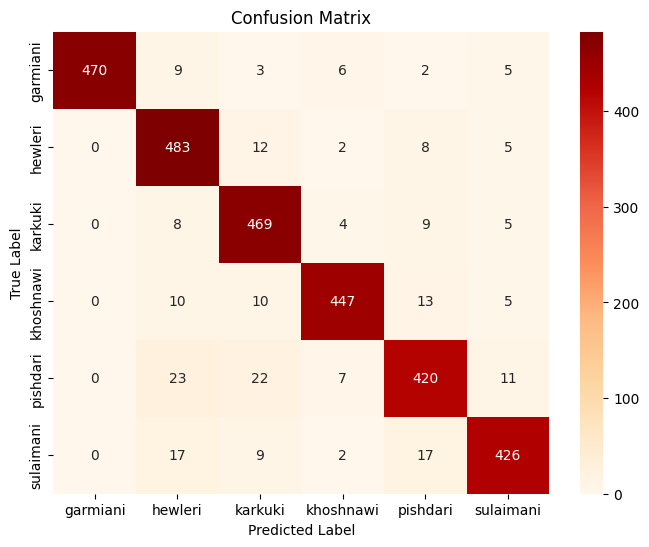}
		\caption*{(b) Confusion matrix of CNN model}
	\end{minipage}
	
	\vspace{\floatsep} 
	
	\begin{minipage}{\textwidth}
		\centering
		\includegraphics[width=0.5\linewidth]{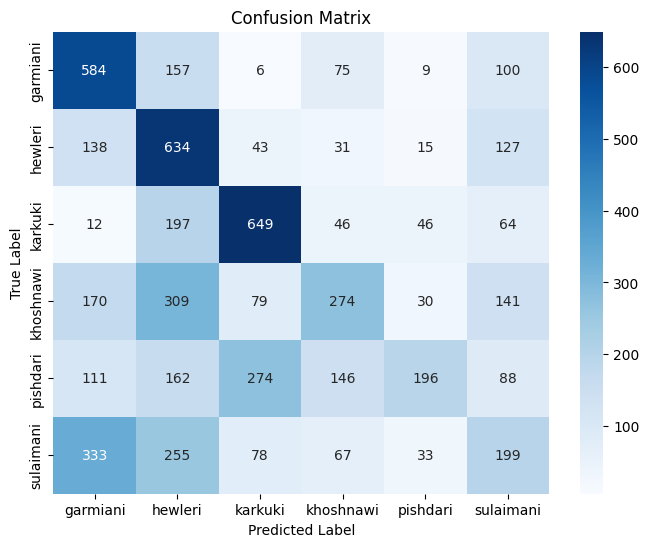}
		\caption*{(c)Confusion matrix of RNN-LSTM model}
	\end{minipage}
	
	\caption{The confusion matrix of the three models}
	\label{fig:model_con_mat}
\end{figure}

While we examine the classification metrics in Table \ref{Classification metrics}, the validation in Table \ref{Detecting single tracks from each subdialect result}, and the confusion matrix which is provided in Figure \ref{fig:model_con_mat} for the three models highlight the interrelation between various subdialects. These matrices corroborate with the geographical proximity of the subdialects and the confusion that the system or humans might have in their correct classification.

\begin{table}
	\begin{center}
		\caption{Detecting single tracks from each subdialect result}
		
		\begin{tabular}{|p{4cm}|p{3.33cm}|p{1.85cm}|p{1.85cm}|p{1.85cm}|}
			\hline
			{Transcribed (Kurdish)} & {Transcribed (Latin)} & {Actual subdialect} &  {Machine predicted} &  {Human predicted}
			\\ \hline
			{\small{\RL{‫وە ئەلەرم هەڵەسم‬‬‬}}}& We elerm hellesm &  Garmiani & Garmiani & Garmiani	\\ \hline 
			{\small{\RL{سبەینان هەڵەستم‬}}}& Sbeyinan hellestm &  Karkuki & Pishdari & Pishdari, Karkuki	\\ \hline 
			{\small{\RL{وەڵا بەخۆم هەردەستم‬}}}& Wella bexom herdestm &  Hewlri & Pishdari & Hewleri		\\ \hline 
			{\small{\RL{دوانزەی نیوەڕۆ هەڵەسم‬}}}& Dwanzeyi niywerro hellesm &  Sulaimani & Sulaimani & Sulaimani	\\ \hline
			{\small{\RL{سبەینان بە مۆبایلی هەرەستم}}}& Sbeyinan be mobayl\^i herestm &  Khoshnawi & Khoshnawi & Hawleri, Khoshnawi \\ \hline
			{\small{\RL{دواییش هەڵدەستم‬}}}& {Dway\^i\c{s}} helldestm &  Pishdari & Pishdari & Karkuki, Pishdari	\\ \hline 
		\end{tabular}
		\label{Detecting single tracks from each subdialect result}
	\end{center}
\end{table}

\section{Conclusion}
\label{Sec-Conc}
This thesis focuses on creating a classification system for the Kurdish-Sorani subdialects located in the KRI, specifically for the subdialects of Hewleri, Garmiani, Karkuki, Khoshnawi, Pishdari and Sulaimani. The data collection presented several challenges due to the lack of available social media or references to gather from. However, this thesis successfully collected audio data, making it one of the first datasets to include audio recordings for most of these subdialects. The collected dataset, named Sorani Nas, includes approximately 30 hours of audio data, primarily comprising spontaneous speech from mentioned subdialects.

The thesis comprehensively explains the dataset preparation, including pre-processing steps and dataset balancing techniques. Additionally, it delves into the methodology used for the three models employed in the study: ANN, CNN and RNN-LSTM. This detailed methodology is crucial for understanding the approach taken in the study. Based on a thorough review of existing literature, a model was developed to categorise the Kurdish-Sorani subdialects specifically using the provided dataset. The ANN, CNN and RNN-LSTM models were trained on the same dataset, and different segment durations, dataset balancing techniques, and model parameters were modified to enhance accuracy. Experimental comparisons were conducted among the three models, and CNN and RNN-LSTM were the most recommended models by other researchers. Consequently, a comparison was made between these two models while also including ANN for evaluation purposes. The experimental results demonstrate that the ANN and CNN approaches are efficient, effective and reliable, achieving accuracies  of 93\% and 96\%, respectively, across all classes. Furthermore, the proposed technique outperforms existing algorithms regarding both speed and accuracy.

Overall, this thesis significantly contributes to classifying Kurdish-Sorani subdialects in the KRI region. The developed models and dataset provide valuable resources for further research in this domain, and the findings highlight the efficacy and reliability of the proposed approach in accurately categorising these subdialects.

In the future, we are interested in the further enrichment and extension of the scope of the subdialects to include other Sorani subdialects that are found in Iran. Expanding to encompass other dialects, like Kurmanji and Hawrami subdialects, would also be a valuable direction for future research. Also, transcription of the dataset could hep other types of research on Sorani subdialect studies.

\section*{Online Resources}

The dataset is partially publicly available for non-commercial use under the CC BY-NC-SA 4.0 license 5 at \url{https://github.com/KurdishBLARK/Sorani Nas}.

\section*{Acknowledgments}
We acknowledge and appreciate very much the the value of the contribution that the participants in the data collection process. We pay our utmost respect to every one of them. Their names become available along with Sorani Nas dataset. The dataset is partially available now and will be fully available upon the publication of the peer reviewed version of the paper.

\bibliographystyle{lrec}
\bibliography{Ku-SubDialRecognition}

\end{document}